\documentclass[preprint,authoryear,11pt]{elsarticle}
\usepackage{geometry}
\usepackage{dsfont}
\usepackage{amsmath}
\usepackage{cancel}	
\usepackage{amssymb}
\usepackage{amsbsy}
\usepackage{amsthm}
\usepackage{amsfonts}
\usepackage{latexsym}
\usepackage{bm}
\usepackage{verbatim}
\usepackage{mathrsfs}
\usepackage{graphicx}
\usepackage{subfigure}
\usepackage{wrapfig}
\usepackage{fmtcount}
\usepackage{color}
\usepackage{booktabs}
\usepackage{lscape}
\usepackage{mathrsfs}
\usepackage{makecell}
\usepackage{lineno,hyperref}
\usepackage{graphicx}
\usepackage{multirow}
\usepackage{comment}
\usepackage{chngcntr}
\usepackage{tcolorbox}
\usepackage{booktabs}
\usepackage{longtable}
\tcbuselibrary{skins, breakable, theorems}
\usepackage{lipsum}
\usepackage[normalem]{ulem}
\journal{CMAME}
\begin{document}

\begin{frontmatter}

\title{A mechanistic-based data-driven approach to accelerate structural topology optimization through finite element convolutional neural network (FE-CNN)}

\author[dtua]{Tianle Yue}
\author[dtua]{Hang Yang}
\author[dtua,dtub,ningbo]{Zongliang Du}
\author[dtua,dtub,ningbo]{Chang Liu}
\author[aucegupt]{Khalil I. Elkhodary}
\author[dtua,dtub,ningbo]{Shan Tang\corref{cor1}}
\cortext[cor1]{Corresponding author: shantang@dlut.edu.cn}
\author[dtua,dtub,ningbo]{Xu Guo\corref{cor2}}
\cortext[cor2]{Corresponding author too: guoxu@dlut.edu.cn}

\address[dtua]{State Key Laboratory of Structural Analysis for Industrial Equipment, Department of Engineering Mechanics, Dalian University of Technology, Dalian, 116023, PR China}
\address[dtub]{International Research Center for Computational Mechanics, Dalian University of Technology, Dalian, 116023, PR China}
\address[ningbo]{Ningbo Institute of Dalian University of Technology, Ningbo 315016, China}
\address[aucegupt]{The Department of Mechanical	Engineering, The American University in Cairo, 11835, New Cairo, Egypt}

\begin{abstract}
In this paper, a mechanistic data-driven approach is proposed to accelerate structural topology optimization, employing an in-house developed finite element convolutional neural network (FE-CNN). Our approach can be divided into two stages: offline training, and online optimization. During offline training, a mapping function is built between high and low resolution representations of a given design domain. The mapping is expressed by a FE-CNN, which targets a common objective function value (e.g., structural compliance) across design domains of differing resolutions.  During online optimization, an arbitrary design domain of high resolution is reduced to low resolution through the trained mapping function. The original high-resolution domain is thus designed by computations performed on only the low-resolution version, followed by an inverse mapping back to the high-resolution domain. Numerical examples demonstrate that this approach can accelerate optimization by up to an order of magnitude in computational time. Our proposed approach therefore shows great potential to overcome the curse-of-dimensionality incurred by density-based structural topology optimization. The limitation of our present approach is also discussed.

\end{abstract}
\end{frontmatter}


\section{Introduction}\label{sec1}
Structural topology optimization seeks to distribute a given amount of material inside a design domain, such that the optimized structure would exhibit maximal performance under prescribed conditions. Such optimization demonstrates great potential for the creation of innovative structural designs, without relying on extensive a priori knowledge. Due to its importance in engineering applications, it has attracted and sustained research attention for a long time, inspired by seminal works such as, \citep{prager1977optimal,cheng1981investigation,bendsoe1988generating,zhou1991coc}. Thorough reviews of state-of-art techniques in structural topology optimization have been also presented in, \citep{rozvany2009critical,guo2010recent,maute2013topology,deaton2014survey}. For practical engineering, however, the computational cost of structural topology optimization may be prohibitive, even with a single loading case, especially with the use of the classical SIMP method (solid isotropic material with penalization), cf.   \citep{bendsoe1989optimal,rozvany1989continuum}. In the SIMP method, the number of design variables scales directly with the total number of elements involved in the finite element analysis (FEA). An explosion of design variables, and of computational cost, quickly become inescapable.

With the fast growing developments in artificial intelligence (AI), machine learning (especially deep learning) is becoming an increasingly relevant tool to improve on these traditional techniques of structural topology optimization. For example, \cite{ulu2016data} developed a data-driven approach to estimate optimized topologies under various loading cases. In the same vein, \cite{sosnovik2019neural} and \cite{banga20183d} adopted convolutional neural networks (CNNs) to build a relation between the near-optimal results obtained from the early iterative steps and the final optimal structure, based on SIMP. Furthermore, \cite{Yu} sought a near-optimal topological design, without resorting to any iteration, through a combination of CNN and generative adversarial networks (GANs). That is, one would first input the boundary conditions of a topology optimization problem, and then generate an optimal structure with fewer pixels through CNN. The structure is then refined through GAN to obtain a higher-resolution optimal structure. Conversely, \cite{Lee} and \cite{Takahashi} trained a relationship between all design variables and structural compliance (or its sensitivity) through CNN, as such replacing the time-consuming FEA steps in optimization. Different from the foregoing approaches, \cite{lei2019machine} targeted real-time structural topology optimization by training with supported vector regression (SVR) and K-nearest-neighbors (KNN) on the results obtained from an explicit topology optimization framework that is based on \textit{moving morphable components} (MMC)   \citep{guo2014doing,guo2016explicit,zhang2016new}. The training process takes much less computational time because MMC needs much fewer design variables.

Within all aforementioned AI-aided structural topology optimization methods, the machine learning tools (e.g., CNN, GAN, SVR or KNN) are employed as black boxes that establish a desired mapping, without need for a prior knowledge of the mechanics. These methods have a high demand for data, and offer relatively low accuracy and transferability. Hence, minor changes to the boundary conditions, or to the design domain, would require regenerating these large datasets anew, and retraining, which seriously limits the utility of all these methods. As a recent exception, \citep{chi2021universal} adopted deep neural networks (DNNs) that are trained during (rather than before)  topology optimization, thus avoiding poor transferability.

In this work we pursue a different path, one inspired by the multi-resolution method that was used to improve the efficiency of topology optimization \citep{nguyen2010computational,liu2018efficient}. We start the structural optimization process on a design domain with many elements (design variables), labeling it as the larger domain. During optimization, a CNN is modified to map this larger domain (gray image) to an image with fewer pixels (labelled as the smaller domain). The mapping function is only related to the gray level of the image, not to the boundary conditions or to the prescribed load. Then, a finite element computation is performed, replacing the fully connected layers of conventional CNNs, to compute the compliance of the smaller domain. Our modified CNN is then trained to realize the same compliance for the larger and the smaller domains, under prescribed boundary conditions and loads. Once the network has been trained, it can be used to carry out optimization on a much larger design domain. During optimization, FEA is always carried out on the smaller domain only, and the gray levels (design variables) of the larger domain are updated to obtain the optimal structure using a back-propagation algorithm applied to the trained CNN. Numerical examples that we solved confirm that our approach can perform fast optimization on large design domains, and can significantly overcome the curse of dimensionality.

Our paper is organized as follows. In Section 2 the classical topology optimization method of SIMP is overviewed, since it will embed our proposed approach. In Section 3 our proposed approach and in-house developed FE-CNN are outlined. More details about FE-CNN are then discussed in Section 4. Section 5 presents our key numerical examples. Concluding remarks are then offered in Section 6.

\section{Topology optimization problem}
In this paper, our AI-based structural topology optimization adopts the SIMP method in which the design variables can be thought as of as the gray level of a pixel/voxel. We work with the FEA representation of the minimum compliance of a designed structure. The problem is stated as follows, which aims to minimize the compliance of a structure under constraints,
 \begin{equation} \label{eq1}
 \left\{
 \begin{aligned}
\min\limits_{\bm{x}} \quad &C(\bm{x})=\bm{U}^T\bm{KU}=\sum_{e=1}^{n}(x_e)^p\bm{u}_e^T\bm{k}_{0}\bm{u}_e \\
s.t. \quad &\frac{V(\bm{x})}{V_0} \leq f \\
  &\bm{KU}=\bm{F} \\
  &0 < x_{\text{min}} \leq min\bm{(x)} \leq max\bm{(x)} \leq 1
\end{aligned}
\right.
\end{equation}
where $C$ represents the compliance; $\bm{U}$, $\bm{F}$ and $\bm{K}$ indicate the global displacement vector, force vector, and stiffness matrix respectively. $\bm{u}_e$ and $\bm{k}_{0}$ denote displacement vector and elemental stiffness matrix of solid element, respectively. In the objective function, $n$ is the total number of elements used to discretize the design domain; the superscript $p$ indicates the penalization power index. A typical value of $3$ is  adopted, as in previous work \citep{sigmund200199}. The design variables $\bm{x}$ are continuous. Usually, the total number of elements in the finite element mesh is the same as the number of design variables. Thus, the length of the design variables vector $\bm{x}$ is $n$. The symbol $x_{\text{min}}$ is the lower bound of $\bm{x}$, introduced to avoid a singularity of the stiffness matrix during the optimization process. In typical applications, $x_{\text{min}}=10^{-3}$ is used. The design variables $\bm{x}$ can represent the vector of pixels or voxels of an image. $V(\bm{x})$ and $V_0$ in the constraints are the material volume, and the design domain volume, respectively. The ratio $f$ between them is the volume fraction, which is specified a priori.

The sensitivity of the objective function ${\partial C}/{\partial x_e}$ is required to solve the optimization problem shown in Eq. (\ref{eq1}), which can be found as
\begin{equation} \label{eq4}
\frac{\partial C}{\partial x_e} = -p(x_e)^{p-1}\bm{u}_e^T\bm{k}_0\bm{u}_e.
\end{equation}
FEA is often used to compute the sensitivities of the objective functions based on this equation. For a very large number of design variables, it is expected that a final shape of fine, complex and detailed geometry will be sought.  This expectation implies that FEA models with highly refined meshes are required, whose  computational cost would be correspondingly high.
\begin{figure}
	\centering
	\includegraphics[scale=0.7]{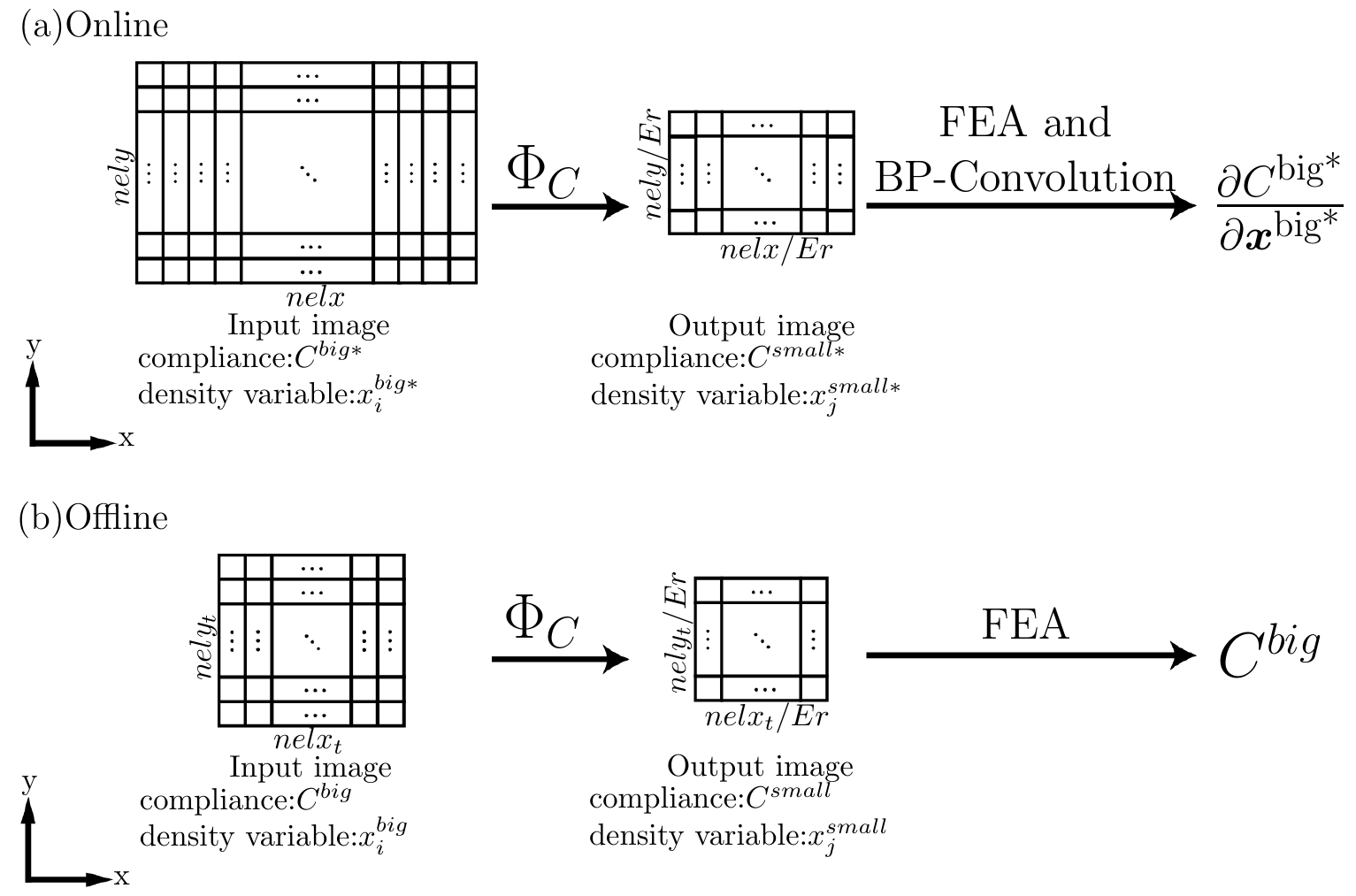}
	\caption{The proposed approach for structural topology Optimization is divided into two stages. (a) Online optimization stage. A mapping function $\Phi_C$ is used to obtained the compliance and sensitivity for the input image with $nelx \times nely$ elements by the computation on the output image with $nelx/Er \times nely/Er$ elements, where the scaling factor $Er$ is integer and greater than $1$. (b) Offline training stage. The mapping function $\Phi_C$ is trained by minimizing the error for the compliance between the output image with $nelx_t/Er \times nely_t/Er$ elements and input image with $nelx_t \times nely_t$ elements. The scale factor is the same as that in the online stage. } \label{fig1}
\end{figure}

\section{FE-CNN model for topology optimization}

We carry out a topology optimization that seeks the minimum structural compliance over a given design domain, as shown to the left of Fig. \ref{fig1} (a). For this 2D problem the design domain is meshed by $nelx$ and $nely$ elements along the $x$ and $y$ directions respectively. Design variables are denoted as $\bm{x}^{\text{big*}}$ and the total number of design variables is $nelx \times nely$. The compliance is denoted as $C^{\text{big*}}$. As mentioned in Section 2, the sensitivities of the compliance with respect to the design variables play a critical role in topology optimization. If FEA is used to compute these sensitivities, as is done per traditional methods, the computational cost would increase sharply, scaling with $nelx$ and $nely$. Instead, to obtain these sensitivities for large-scale topology optimization problems quickly and accurately, a modified CNN may be designed, which we term FE-CNN. Unlike conventional CNN, the architecture of FE-CNN is redesigned based on a priori mechanistic knowledge. Through our in-house developed FE-CNN, the design domain with more elements can be reduced to a mesh of fewer elements ($nelx/Er \times nely/Er$, where the scaling integer $Er$ is greater than $1$), as shown in Fig. \ref{fig1} (a). Both images (scaled and unscaled) possess the same volume fraction, which we achieve by means of the Optimality Criteria (OC) method. This stage is referred to as the OC layer of the CNN. The details of this process are described in Section \ref{sec3.1}. In our proposed approach, FEA needs not be performed on the larger mesh. Rather, our FE-CNN is trained to ensure that the compliances of the larger and smaller meshes are on the par (i.e. $C^{\text{big*}} = C^{\text{small*}}$). The compliance and the sensitivities of the small mesh are computed via FEA. Then, the sensitivities of the larger mesh can be computed as follows,
\begin{equation} \label{eq7}
\frac{\partial C^{\text{big}^{*}}}{\partial x^{\text{big*}}_i} = \frac{\partial C^{\text{small*}}}{\partial x^{\text{big*}}_i} = \sum_{j=1}^{nelx \times nely/r^2}\frac{\partial C^{\text{small*}}}{\partial x^{\text{small*}}_j}\frac{\partial x^{\text{small*}}_j}{\partial x^{\text{big*}}_i},
\end{equation}
where ${\partial C^{\text{small*}}}/{\partial x^{\text{small*}}_j}$ are the sensitivities of the small mesh, which represent the influence of the change of gray levels on the compliance for the small mesh. ${\partial x^{\text{small*}}_j}/{\partial x^{\text{big*}}_i}$ represents the transition function of the density variables between the larger mesh and the smaller mesh. It can be computed by a back propagation convolution (BP-Conv) layer, similar to CNN. The details of this process are described in Section \ref{sec3.2}, after having introduced the method of training for FE-CNN in Section \ref{sec3.1}. In particular, FE-CNN with a scaling factor of $Er=2$ is used for illustration these sections. We also extend to networks with scaling factors $Er=4$ and $Er=8$ using numerical examples, which are discussed in Section \ref{secNu}.

\begin{figure}
	\centering
	\includegraphics[scale=0.8]{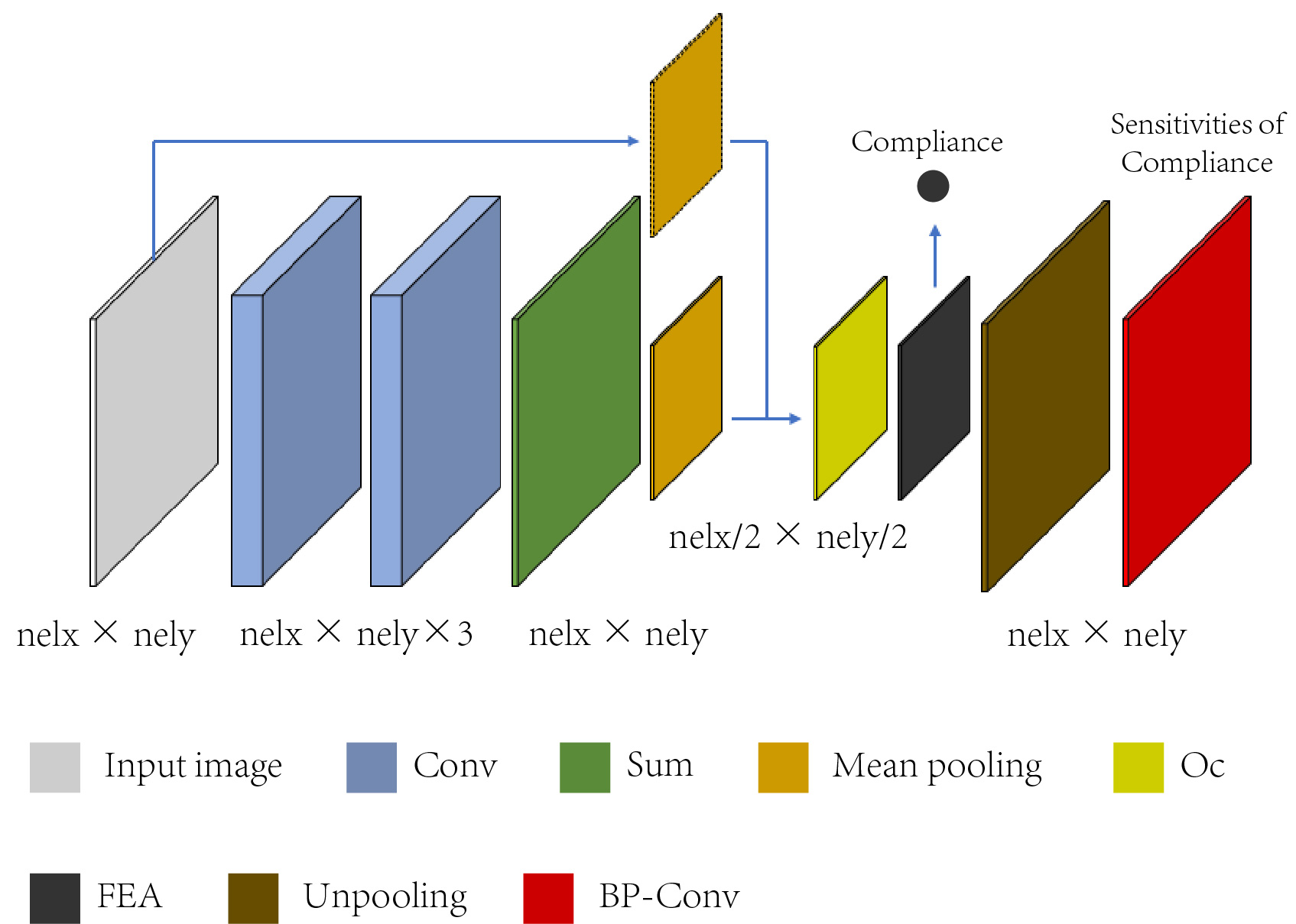}
	\caption{The architecture of the proposed FE-CNN with scaling factor $Er=2$ for structural topology optimization.} \label{fig2}
\end{figure}

In our approach, the larger mesh can be regarded as a high-resolution image, and the smaller mesh can be thought of as a low-resolution image. Our in-house developed FE-CNN serves as a tool to map between these two meshes, while preserving structural compliance under the same loads and constraints. The mapping function between the two images is defined as $\Phi_{\text{C}}$, which can be thought as a function for image compression. Because this mapping function is fully defined by the gray levels of the pixels (not by the loads or boundary conditions), it is reasonable to assume that this mapping function is invariant for all transformations of the same-scale (large:small ratio). As a result, in Fig. \ref{fig1} (b) a training mesh with $nelx_t$ and $nely_t$ elements is shown. Note that $nelx_t\le nelx$ and $nely_t\le nely$. The training mesh can be then reduced to a smaller mesh, having $nelx_t/Er$ and $nely_t/Er$ elements along $x$ and $y$ direction respectively. With the scaling factor $Er$ taken as the same as for the online stage, the mapping function $\Phi_{\text{C}}$ holds for the offline stage (training) as for the online one (optimization), as indicated in Fig. \ref{fig1}. In short, our training mesh with $nelx_t$ and $nely_t$ elements can thus be used to train the image mapping function $\Phi_{C}$ during the offline stage. The resulting trained $\Phi_{C}$ is then used for the online topology optimization problem. We note that $nelx_t/Er$ and $nely_t/Er$ must be integers. 

In summary, as shown in Fig. \ref{fig2} for a scaling factor $Er=2$, our in-house developed FE-CNN is composed of the following: a convolution layer, a pooling layer, a summation layer, an OC layer, an FEA layer, an unpooling layer, and finally a BP-Conv layer. The following sequence is observed: the input image is divided into three channels, each passing through two convolution layers, a summation layer, a pool layer and an OC layer. Then, the compliance and the sensitivities are computed by FEA. Finally, the sensitivities of the small mesh pass through the unpooling layer, and the BP-Conv layer to obtain the sensitivities of the input image.

 \begin{figure}
	\centering
	\includegraphics[scale=0.8]{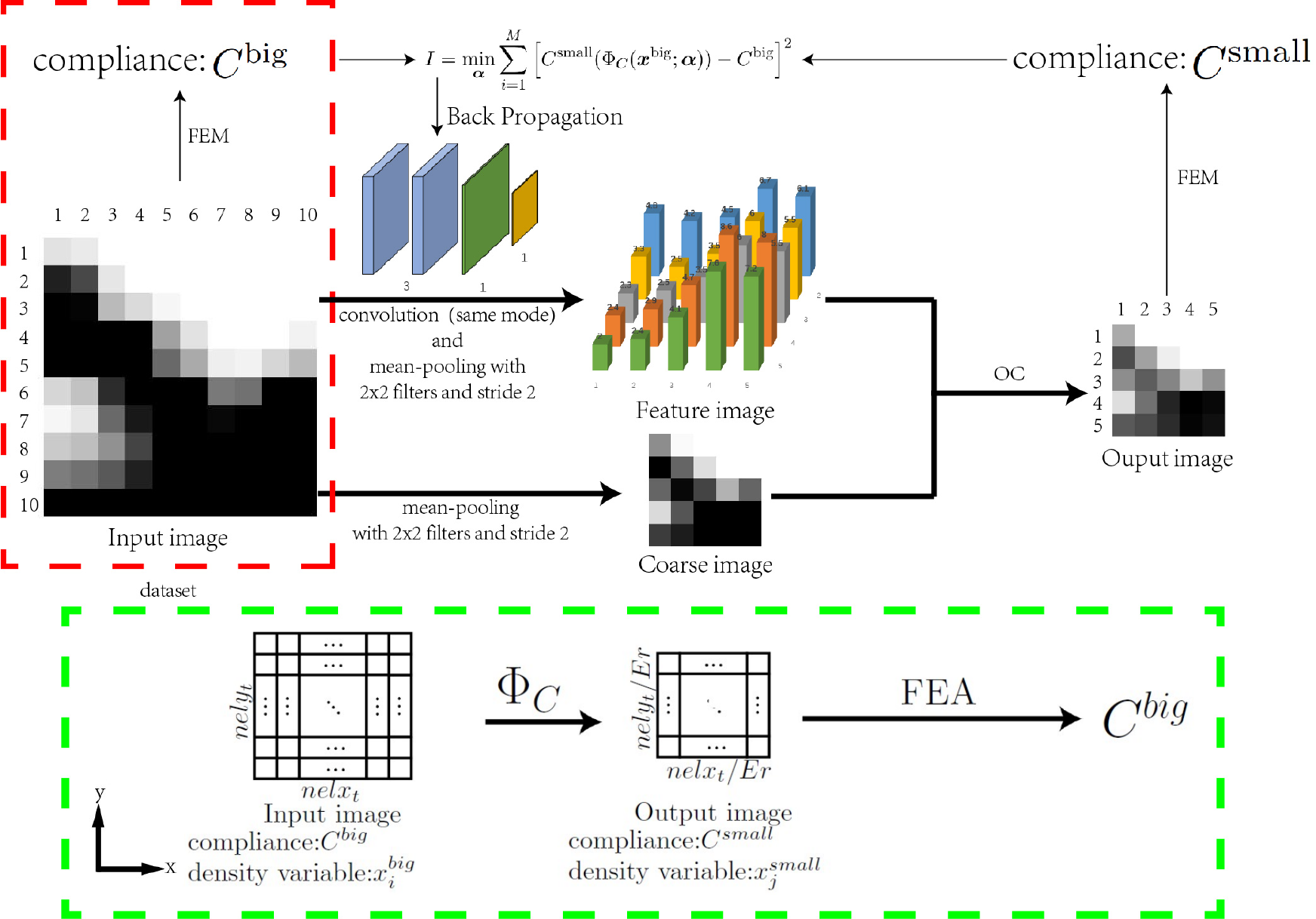}
	\caption{The flowchart of the Offline training, which is the exact realization shown in Fig. 1. The target is to minimize the error between the compliance of output image and the compliance of input image. The compliance of output image is computed by the FE-CNN.} \label{fig3}
\end{figure}

\subsection{Offline training} \label{sec3.1}
This section focuses on training $\Phi_{C}$ offline. Fig. \ref{fig3} shows the detailed flowchart for this training. Input training images (i.e., meshes with design variables $\bm{x}^{\text{big}}$) are taken from a pre-generated dataset. The generation of that dataset (series of large training images) is based on structural topology optimization by SIMP, as is elaborated in Section \ref{sec4.1}. The compliances corresponding to the large input mesh and to the small output mesh are herein designated as $C^{\text{big}}_{i_{s}}$ and $C^{\text{small}}$, respectively. Our training for $\Phi_{C}$ aims to minimize the difference between these two compliances,
\begin{equation} \label{eq5}
I = \min_{\bm{\alpha}}\sum_{{i_{s}}=1}^{M}{\left[C^{\text{small}} (\Phi_{C}(\bm{x}_{i_{s}}^{\text{big}};\bm{\alpha});...) - C_{i_{s}}^{\text{big}}\right]}^2.
\end{equation}
We assume the same loads and constraints hold for the meshes in this equation, where $M$ represents the number of samples in the dataset, and the subscript $i_{s}$ the sample number. The compliance of the small mesh depends on the parameters of FE-CNN (weights and bias $\bm{\alpha}$), and on $\bm{x}^{\text{big}}$.

To proceed with the minimization of this function, $C^{\text{small}}$ needs to be computed using $\bm{x}^{\text{small}}$. To obtain $\bm{x}^{\text{small}}$ we first generate a feature image via a convolution layer and a mean-pooling layer, to obtain $\bm{x}^{\text{feature}}$. In this step, the input image is separated into three channels according to the design variables $\bm{x}^{\text{big}}$. Then, a coarse image is generated through mean-pooling, to obtain $\bm{x}^{\text{coarse}}$. Lastly, the output image with the small mesh combines the feature image and coarse image using the OC method as,

\begin{equation} \label{eq8}
x_i^{\text{small}}=\left\{
\begin{aligned}
&x^{\text{small}}_{\text{min}} \quad & & \quad x_i^{\text{coarse}}\sqrt{\frac{x_i^{\text{feature}}}{\lambda}} \leq x^{\text{small}}_{\text{min}} \\
&x_i^{\text{coarse}}\sqrt{\frac{x_i^{\text{feature}}}{\lambda}}  & & \quad x^{\text{small}}_{\text{min}}<x_i^{\text{coarse}}\sqrt{\frac{x_i^{\text{feature}}}{\lambda}}<1 \\
&1 & & \quad 1 \leq x_i^{\text{coarse}}\sqrt{\frac{x_i^{\text{feature}}}{\lambda}}
\end{aligned}
\right.,
\end{equation}
where $x^{\text{small}}_{\text{min}}$ are lower bounds on densities. In typical applications $x^{\text{small}}_{\text{min}}=10^{-3}$. The symbol $\lambda$ denotes the Lagrange multiplier used to satisfy the volume constraint, and is solved for by a bisection algorithm. 

\begin{figure}
\centering
\includegraphics[scale=0.7]{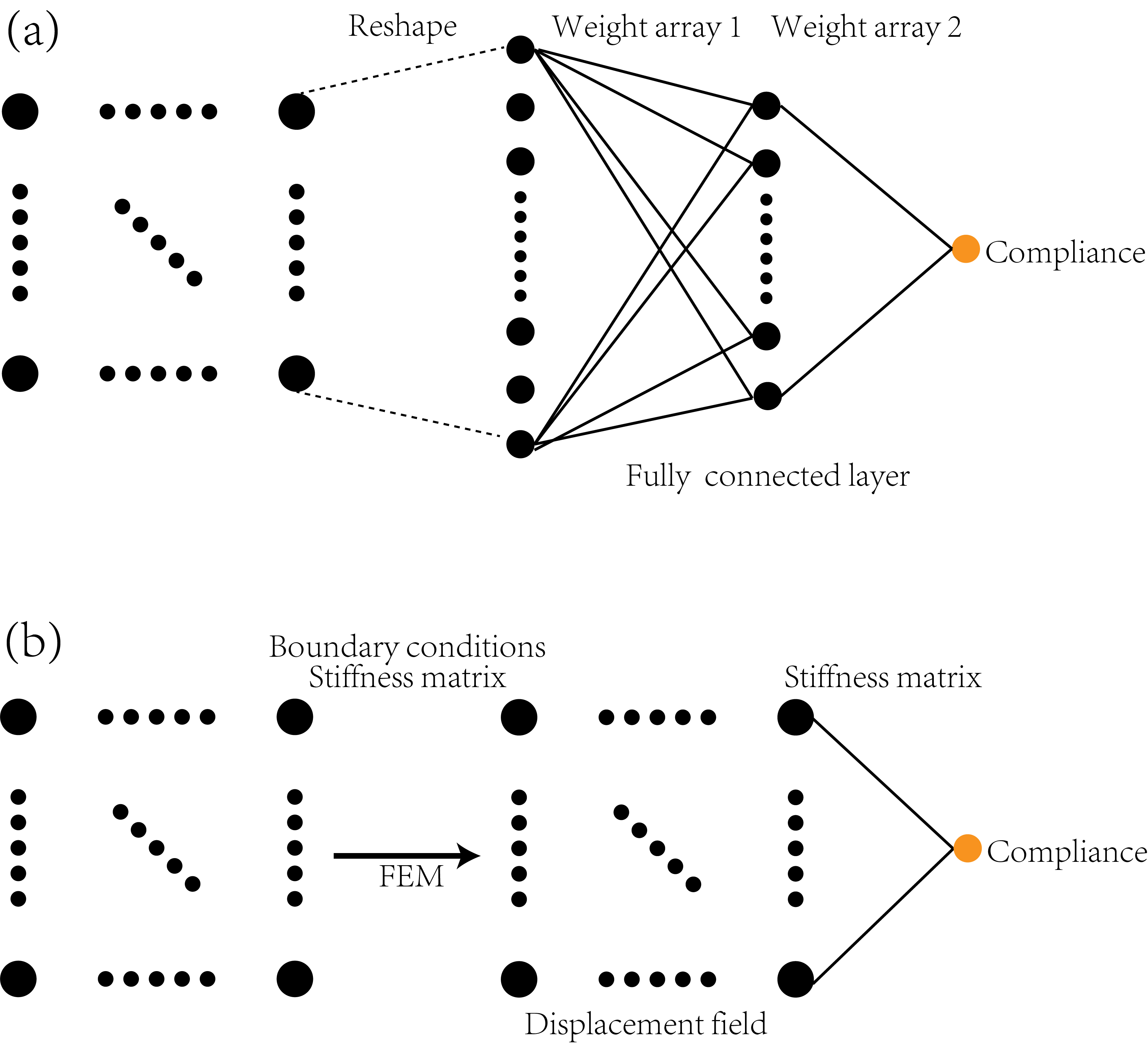}
\caption{Comparison between the fully connection neural network in conventional CNNs and FEA layer in our proposed FE-CNN. (a) fully connected layer. (b) FEA layer.} \label{fig5}
\end{figure} 

Through this outlined procedure, an output image can be obtained with the same volume fraction $f$ as for the input image. This OC layer in our FE-CNN can be compared to the first fully connected layer in conventional CNN, which  processes the feature image. With a known $\bm{x}^{\text{small}}$, we then compute by FEA the compliance for the output image and the error function in Eq. (\ref{eq5}). By contrast, conventional CNN, e.g. \cite{Lee}, uses a fully connected layer to obtain the compliance (without resorting to FEA). Fig. \ref{fig5} highlights how our approach differs.  In conventional CNN, convolution and pooling are used to compute the compliance. Instead, our FEA layer leverages the rules of mechanical analysis to reduce the number of the parameters involved in the fully connected layer of conventional CNN, affording  lesser training time and higher accuracy for our FE-CNN.

For the purpose of online optimization, the partial derivatives of the compliance of the smaller mesh with respect to CNN parameters are needed to compute the sensitivities, 
\begin{equation}\label{bp}
\frac{\partial C^{\text{small}}}{\partial \bm{\alpha}} = \frac{\partial C^{\text{small}}}{\partial x_i^{\text{small}}}\frac{\partial x_i^{\text{small}}}{\partial x_j^{\text{feature}}}\frac{\partial x_j^{\text{feature}}}{\partial \bm{\alpha}},
\end{equation}
where for ${\partial C^{\text{small}}}/{\partial x_i^{\text{small}}}$ we refer to Eq. (\ref{eq4}), and ${\partial x_i^{\text{small}}}/{\partial x_j^{\text{feature}}}$ can be derived from Eq. (\ref{eq8}) as,
\begin{equation} \label{eq9}
\frac{\partial x_i^{\text{small}}}{\partial x_j^{\text{feature}}}=\left\{
\begin{aligned}
&\frac{x_j^{\text{coarse}}}{2\lambda}\sqrt{\frac{\lambda}{x_j^{\text{feature}}}} &\ &i = j\\
&0           &\ &i \neq j
\end{aligned}
\right.,
\end{equation}
while ${\partial x_j^{\text{feature}}}/{\partial \bm{\alpha}}$ can be obtained through the back propagation algorithm of CNN \citep{bouvrie2006notes}.
\begin{figure}
	\centering
	\includegraphics[scale=0.8]{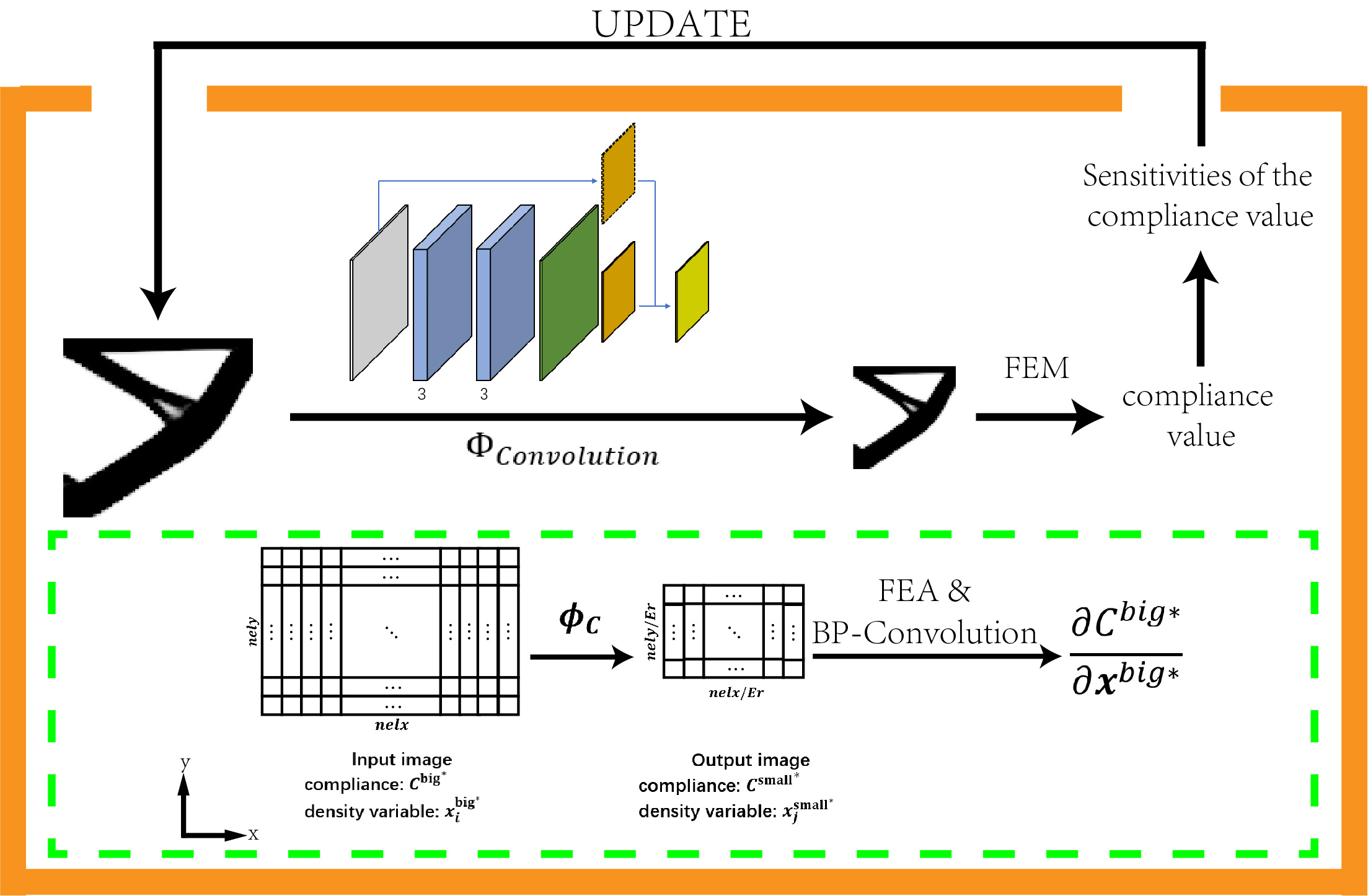}
	\caption{The overall flowchart of the online optimization. The trained FE-CNN is used to accelerate the structural topology optimization by reducing the size of the mesh and the related computational cost.} \label{fig6}
\end{figure}

\subsection{Online optimization} \label{sec3.2}
Let us return to the online optimization problem shown in Fig. \ref{fig6}. The trained $\Phi_{C}$ can now be used with this problem. The sizes of the input image $nelx \times nely$ are determined based on user application.  $\Phi_{C}$ is thus used to generate output images of size $nelx/Er \times nely/Er$. The scaling factor $Er$ is the same as that used for the offline stage. Topology optimization by the SIMP method requires both the compliance and the sensitivities of the compliance for the larger mesh to be available. We here obtain them through a smaller mesh. Design variables $\bm{x}^{\text{big*}}$ for a previous iteration on the larger mesh, with size $N = nelx \times nely$, are fed into the trained FE-CNN. By taking $Er=2$, a scaled down image with size $nelx/2 \times nely/2$ is generated. Its compliance $C^{\text{small*}}$ and  sensitivities in terms of $\bm{x}^{\text{small*}}$. (i.e. $\partial {C^{\text{small*}}}/{\partial x^{\text{small*}}_j} $) are quickly solved for on the corresponding smaller mesh. Next, the compliance and the sensitivities for the larger mesh are computed. We require that $C^{\text{big*}} = C^{\text{small*}} $ after image transformation. The sensitivities for the larger mesh can be obtained through Eq. (\ref{eq7}) where,
\begin{equation} \label{eq10}
\frac{\partial x^{\text{small*}}_j}{\partial x^{\text{big*}}_i} = \frac{\partial x_j^{\text{coarse}}}{\partial x_i^{\text{big*}}}\sqrt{\frac{x_j^{\text{feature}}}{\lambda}} +  \frac{x_j^{\text{coarse}}}{2\lambda}\sqrt{\frac{\lambda}{x_j^{\text{feature}}}}\frac{\partial x^{\text{feature}}_j}{\partial x^{\text{big*}}_i},
\end{equation}
and
\begin{equation} \label{eq11}
\frac{\partial x_j^{\text{coarse}}}{\partial x_i^{\text{big}}}=\left\{
\begin{aligned}
&\frac{1}{4} &\ &4(i-1) \leq j \leq 4i\\
&0           &\ &\text{otherwise}
\end{aligned}
\right.,
\end{equation}
and ${\partial x^{\text{feature}}_j}/{\partial x^{\text{big}}_i}$ being computed via the back propagation algorithm of CNN \citep{bouvrie2006notes}. After the compliance and the sensitivities of the larger mesh image have been evaluated, the OC method outlined in \citep{sigmund200199} is used to update all design variables.  We may then proceed to the following iteration of  optimization. The process is terminated once convergence takes place to within tolerance.

\subsection{FE-CNN vs. multi-resolution topology optimization} \label{sec3.3}

 \begin{figure}
	\centering
	\includegraphics[scale=0.7]{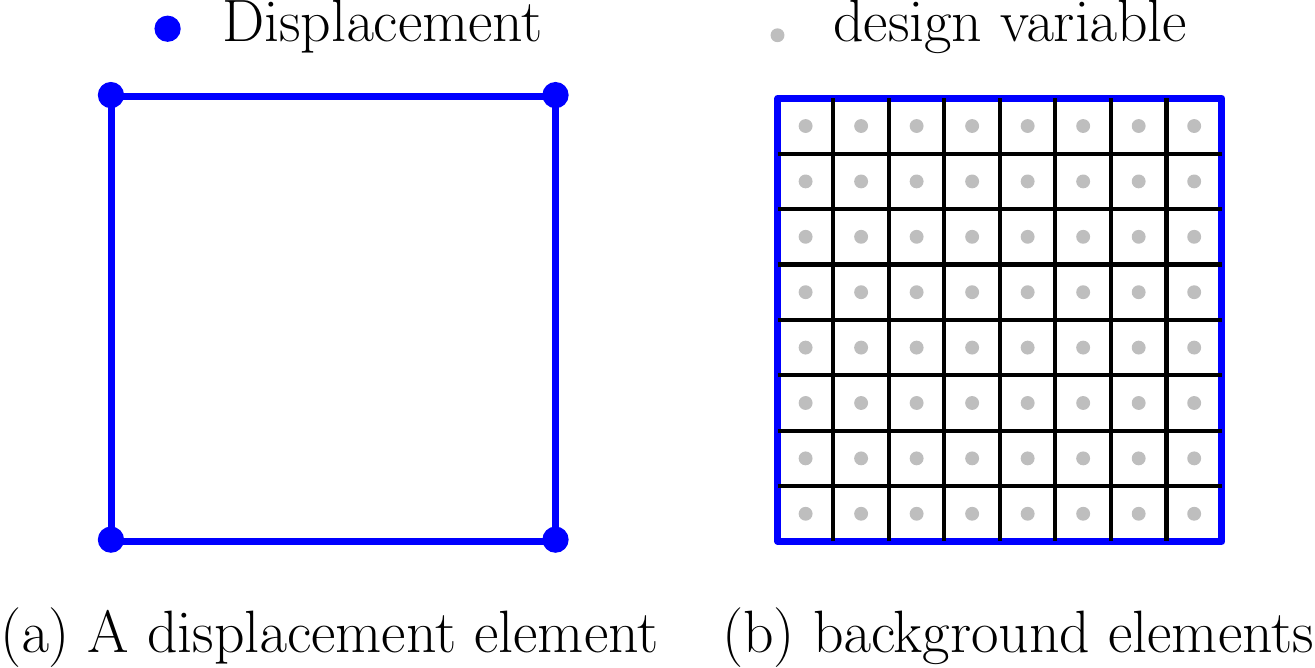}
	\caption{A schematic illustration of (a) a displacement element and (b) background elements.} \label{figMTOP}
\end{figure}

In the Introduction we noted that the Multi-resolution Topology Optimization (MTOP) method can greatly accelerate topology optimization. It can obtain high resolution designs with relatively low computational cost \citep{nguyen2010computational}. Like our proposed approach, a small-scale mesh is generated for FEA in MTOP. Nevertheless, MTOP employs approaches that are different from ours to compute the stiffness matrix, and the sensitivity of the compliance with respect to the design variables. In MTOP, the stiffness matrix of each element in the finite element analysis is computed as follows,
\begin{equation} \label{eqMTOP1}
\bm{K}_e = \int_{\Omega_e} \bm{B}^{T}\bm{DB}\, d\Omega \simeq \sum_{i=1}^{N_n}(x_i)^p\bm{k}_0\mid_{i}A_i,
\end{equation}
where $\bm{K}_e$ is the stiffness matrix of the displacement element $e$, $\bm{B}$ is the strain-displacement matrix of shape function derivatives, $\bm{D}$ is the constitutive matrix, $N_n$ is the number of design variables in the displacement element domain, and $A_i$ is the area of a background element. A displacement element, its design variables and background elements are shown in Fig. \ref{figMTOP}. The sensitivity of the compliance with respect to design variables is computed as follows,
\begin{equation} \label{eqMTOP2}
\frac{\partial C}{\partial x_i} = -\bm{u_e^T}\frac{\partial K_e}{\partial x_i}\bm{u_e}=-p(x_i)^{p-1}\bm{u_e^T}\bm{k}_0\bm{u_e}\mid_{i}A_i,
\end{equation}
where $x_i$ is the design variable in the domain of displacement element $e$. The meaning of other symbols can refer to Eq. \ref{eq1}.

In MTOP, the stiffness matrix of a displacement element is computed using the average of the design variables which are covered by it. The sensitivity of the compliance with respect to each design variable is computed in the background elements, instead of the displacement element, while the sensitivity based on displacements is computed for a single displacement element. In other words, the design domain is divided into several areas to compute stiffness matrices independently. If two design variables are covered by different displacement elements, the sensitivities of the compliance with respect to these design variables are computed twice, no matter how close they are. As a result, the accuracy of the sensitivity analysis is degraded because of mismatches between the displacements and stiffness matrices used in Eq. (\ref{eqMTOP2}). Moreover, a large projection radius is often required to avoid checkerboard patterns and discontinuities in MTOP, losing the finer details in the optimized structure.

Conversely, in our approach, a small-scale image is generated through FE-CNN to compute a stiffness matrix for both FEA and sensitivity analysis, so that the displacements and stiffness matrices in Eq. (\ref{eq4}) and Eq. (\ref{eq7}) are consistent. Meanwhile, the stiffness matrix of an element in our approach is dependent on design variables across a larger area than compared to MTOP. Thus, the sensitivity of the compliance with respect to each design variable is computed based on the displacements of many elements, instead of just one. Compared with MTOP, our sensitivity field is smoother and more representative. Detailed comparisons via numerical examples are presented in Section \ref{MTOPCom1} and Section \ref{MTOPCom2}.

\section{Details of FE-CNN}
In this section we discuss FE-CNN in fuller detail. If the reader wishes to reproduce this work, this section should be read attentively.

\subsection{Dataset} \label{sec4.1}
To train our FE-CNN, a multitude of grayscale images representing the topology of structure and its corresponding boundary conditions are required. These images should also span the stress-strain states that could arise during the process of topology optimization.  These images are generated using the 88-lines of code in \citep{andreassen2011efficient}.

 \begin{figure}
\centering
\includegraphics[scale=0.8]{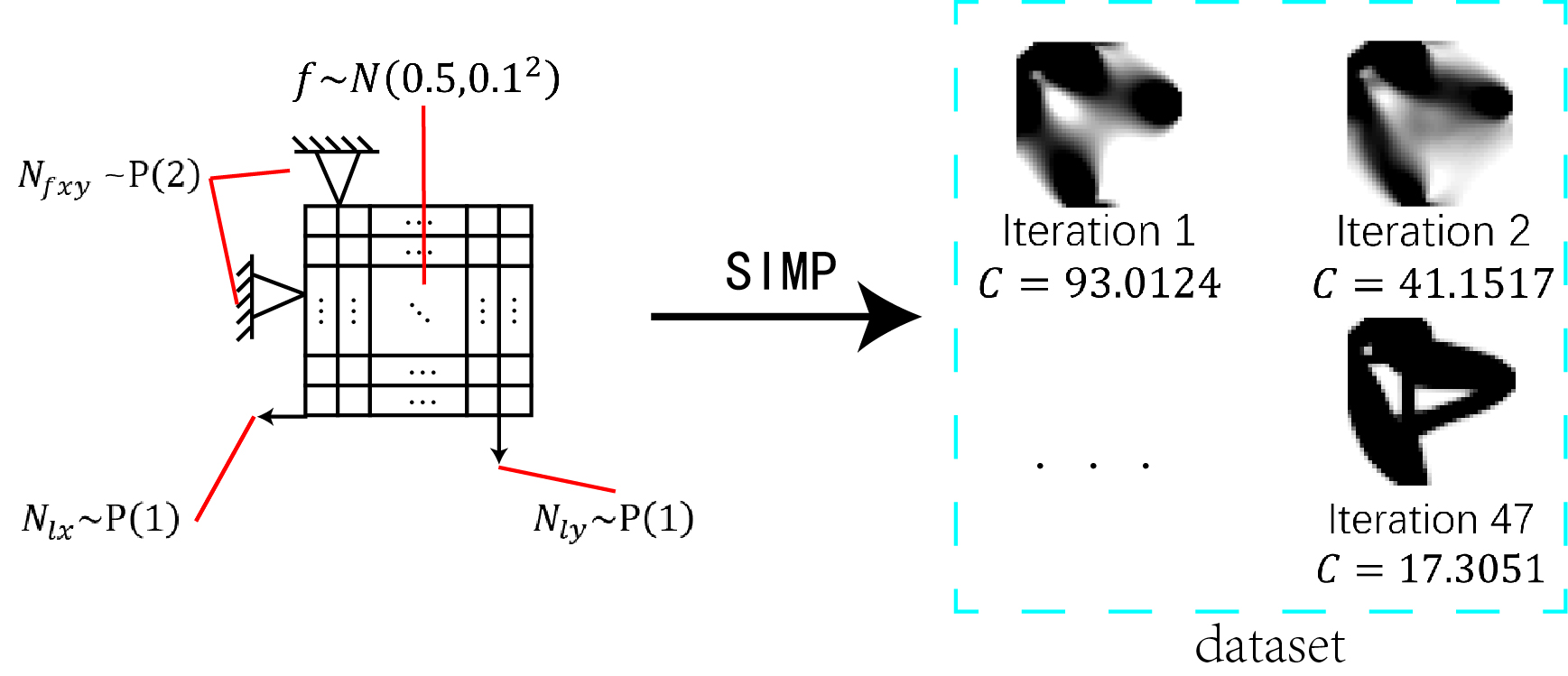}
\caption{A schematic of the dataset generation for offline training. Firstly, a structural topology optimization problem is set. The constraints and loads are imposed randomly. Then the structure is optimized by SIMP method. All the structures computed during each iteration are put into the dataset.} \label{fig7}
\end{figure}

Fig. \ref{fig7} shows the details followed for the generation of a corresponding dataset. To generate the dataset, we focus on the optimization problem of Section 2. The design domain is meshed using $nelx_t\times nely_t$ elements along the $x$ and $y$ directions. We train our FE-CNN with scaling factors of $Er=2$, $Er=4$ and $Er=8$, using a total numbers of elements, $nelx_t = nely_t = 40$, $nelx_t = nely_t = 80$ and $nelx_t = nely_t = 100$, respectively.  For each case, constraints and loads are randomly imposed. The following steps are then followed,

\begin{itemize}
\item Nodes with fixed degrees of freedom in both the $x$ and the $y$ direction, totaling $N_{fxy}$, and the nodes with prescribed forces in the $x$ and $y$ directions, totaling $N_{lx}$ and $N_{ly}$ respectively, are sampled from a Poisson distribution as,

\begin{equation}
N_{fxy} \sim P(2),N_{lx} \sim P(1),N_{ly} \sim P(1),
\end{equation}
as shown in Fig. \ref{fig7}.

\item The positions of these nodes are sampled from a discrete uniform distribution on all boundary nodes, without repetition.

\item All the forces take on a value of $-1$.
\item The volume fraction $f$ is sampled from a normal distribution as,

\begin{equation}
 f \sim N(0.5,0.1^2).
\end{equation}
\end{itemize}

\begin{figure}
\centering
\includegraphics[scale=1.05]{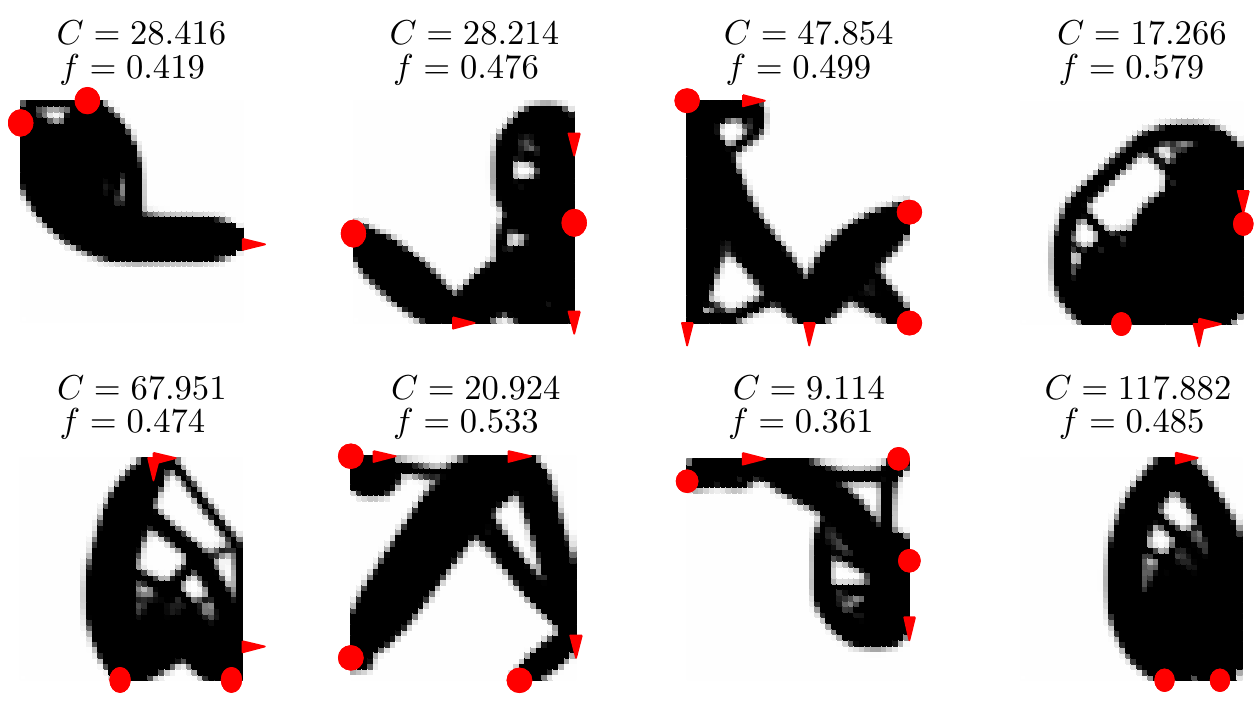}
\caption{Examples of dataset in the offline training. Triangles denote the place with the applied force loads, and circles represent the constraints imposed. The corresponding compliance $C$ and volume fraction $f$ of the structure are also marked.} \label{fig8}
\end{figure}

After the design domain and the boundary conditions are set, we perform 150 iterations of the standard SIMP method, recording the design variables $\bm{x}^{\text{big}}$ and the corresponding compliance for every iteration. The obtained dataset then contains approximately 5,000 samples. Each sample includes a $40 \times 40$ image, a corresponding compliance label, and its boundary conditions. Fig. \ref{fig8} shows some examples for the samples.

\subsection{Convolution layer} \label{sec4.2}
The sizes of the first and the second convolution layer kernels are $3\times3$ and $5\times5$, respectively. The learning rate of the center of the second layer's convolution kernel is $4\times10^{-6}$, and the learning rate for the other parameters is $10^{-4}$.

As shown in Fig. \ref{fig1} there are three channels in the convolution layers in our proposed approach. By adding multiple channels, the network is made more complex, to deal with complex input images. It is important to notice that the input signals of the different channels are expected to be different, otherwise all three channels will yield the same result. The input signals of the three channels are,

\begin{eqnarray}
x_i^{\text{channel-1}} &=&x_i^{\text{big*}}, \\
x_i^{\text{channel-2}} &=&\left\{
\begin{array}{cc}
1 & x_i^{\text{big*}}=1 \\
0 & x_i^{\text{big*}}\neq 1%
\end{array}%
\right.,  \\
x_i^{\text{channel-3}} &=&\left\{
\begin{array}{cc}
0 & x_i^{\text{big*}}=1 \\
x_i^{\text{big*}} & x_i^{\text{big*}}<1%
\end{array}%
\right.,
\end{eqnarray}

\subsection{Pooling layer}
In our proposed approach, all pooling layers obey mean-pooling with $2\times2$ filters and a stride of $2$. Although max-pooling can reduce the value of the error function shown in Eq. (\ref{eq5}), mean-pooling ensures the continuity of the sensitivities shown in Eq. (\ref{eq7}). Moreover, our subsequent use of a BP-Conv layer and an unpooling layer renders this choice of pooling crucially important.

\section{Numerical examples} \label{secNu}
In this section, several examples are investigated to highlight the effectiveness of our proposed approach. Three FE-CNNs with scaling factors $Er=2$, $Er=4$ and $Er=8$ are presented.  The computational time and the value of the optimized objective function are compared with those obtained by efficient implementations of the SIMP method (i.e., 88-lines code in \citep{andreassen2011efficient}). All parameters are fixed across our examples. The Young's modulus and the Poisson's ratio for an isotropic solid material are assumed, with values of $1$ and $0.3$, respectively. The penalty factor $p$ takes on the value of $3$. In addition, all our computations were carried out on a Dell-OptiPlex 7060, with an Intel Core i7-8700, 3.20GHz, CPU, 16GB RAM, and a Windows10 OS. Our computer codes were developed using MATLAB R2020b.

 \subsection{Example 1: A cantilever beam in square design domain}\label{sec5.1}
\begin{figure}
	\centering
	\includegraphics[scale=0.750]{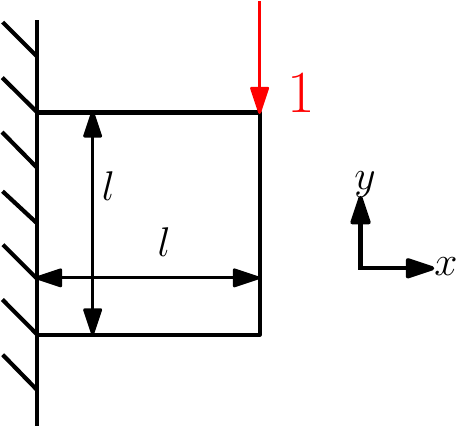}
	\caption{A cantilever beam example with square design domain}
	\label{figc1}
\end{figure}

Topology optimization of a short cantilever beam is examined, as shown in Fig. \ref{figc1}. A square design domain is used; i.e., the same aspect ratio as that of the dataset for FE-CNN training. The cantilever beam is subjected to an end load that is applied at the top corner of the right boundary. 

Table \ref{tab2} summarizes for the different resolutions of FEA used, the optimized design (with $800\times800$ elements), the iteration count, the volume fraction, the projection radius, and the average time cost for the key steps of optimization. That is, the parameters $\bar{t}_{FEA}$, $\bar{t}_{sen}$, and $\bar{t}_{CNN}$ designate the average times of assembling and solving the FE equations, the sensitivity analysis, and FE-CNN respectively. $\bar{t}_{total}$ represents the average time of a full optimization step. The symbols $n_{iter}$, $r$ and $vf$ represent the iteration number at convergence, the projection radius, and the volume fraction, respectively. The quantity $C$ represents the value of the structure's compliance.

\begin{figure}
	\centering
	\includegraphics[scale=1]{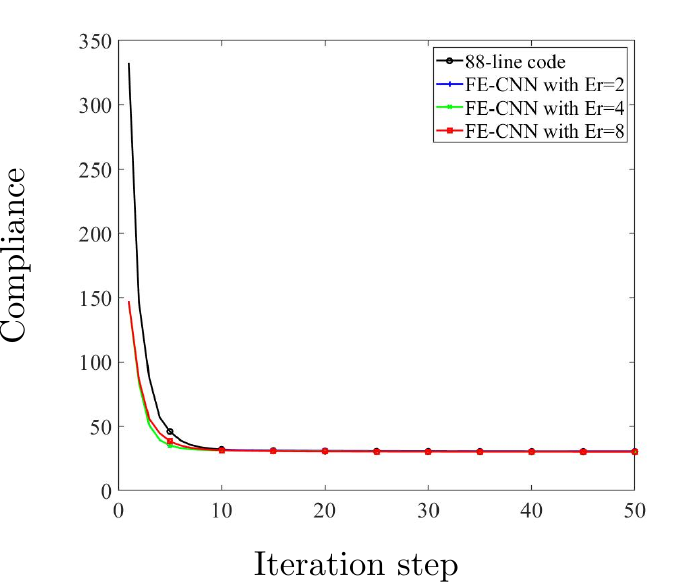}
	\caption{Convergence history after 50 iterations with design variable elements $800\times800$.}
	\label{figl1}
\end{figure}

As shown in Table \ref{tab2}, comparing with the 88-lines of SIMP code, our FE-CNN approach requires only about $27.4\%$, $12.9\%$ and $8.9\%$ of the average time of an iteration, for scaling factors of $Er=2$, $Er=4$ and $Er=8$, respectively. The relative errors of the compliances optimized by the 88-lines of code and our FE-CNN are $0.81\%$, $0.14\%$ and $0.05\%$, respectively. Moreover, compliance convergence histories in our approach and in SIMP are very similar; see Fig. \ref{figl1}.

\begin{table}[htbp]
	\centering
	\caption{Optimization results of the cantilever beam example obtained by 88-lines code and the proposed approach with different FE meshes}
	\begin{tabular}{|c|c|c|c|c|c|c|c|c|c|}
		\hline
		\makecell{Optimized structure\\800x800} & \makecell{Number of\\EF mesh}  & \makecell{$\bar{t}_{FEA}$\\$(s)$} 
		& \makecell{$\bar{t}_{sen}$\\$(s)$} & \makecell{$\bar{t}_{CNN}$\\$(s)$} & \makecell{$\bar{t}_{total}$\\$(s)$} 
		& \makecell{$n_{iter}$} & \makecell{$r$} & \makecell{$vf$} & \makecell{$C$}  \\
		\hline
		\begin{minipage}{0.01\textwidth}	
			\centerline{\includegraphics[width=2.0cm]{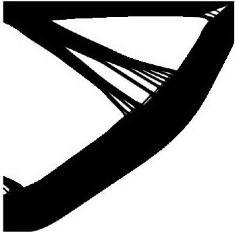}}
		\end{minipage}
		& 800x800 & 8.14  & 0.13  &       & 8.46  & 346   & 2.4   & 0.4   & 30.19 \\
		\hline
		\begin{minipage}{0.01\textwidth}	
			\centerline{\includegraphics[width=2.0cm]{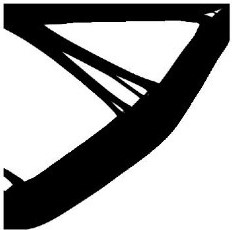}}
		\end{minipage}
		& 400x400 & 1.66  & 0.03  & 0.17  & 2.32  & 604   & 2.4   & 0.4   & 30.43 \\
		\hline
		\begin{minipage}{0.01\textwidth}	
			\centerline{\includegraphics[width=2.0cm]{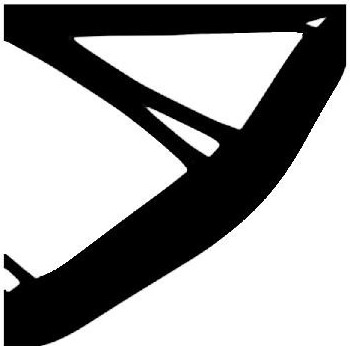}}
		\end{minipage}
		& 200x200 & 0.35  & 0.01  & 0.16  & 1.09  & 668   & 2.4   & 0.4   & 30.23 \\
		\hline
		\begin{minipage}{0.01\textwidth}	
			\centerline{\includegraphics[width=2.0cm]{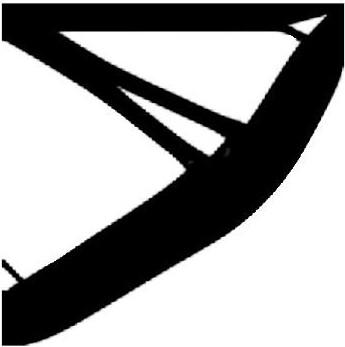}}
		\end{minipage}
		& 100x100 & 0.07  & 0.01   & 0.14  & 0.75   & 525   & 4.8   & 0.4   & 30.17 \\
		\hline
	\end{tabular}%
	\label{tab2}%
\end{table}%

\begin{figure}
	\centering
	\includegraphics[scale=0.750]{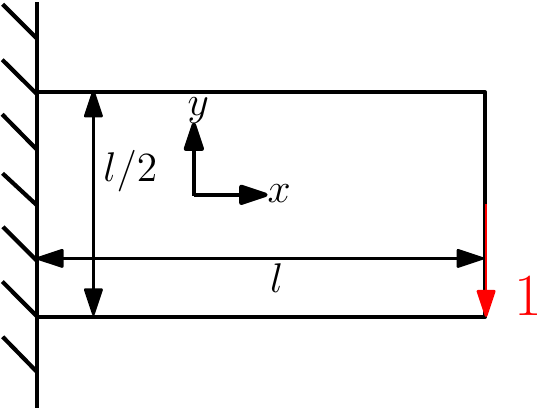}
	\caption{A cantilever beam example with rectangular design domain}
	\label{figc2}
\end{figure}

\subsection{Example 2: A cantilever beam in a rectangular design domain}
A rectangular design domain with an aspect ratio of $2:1$ is used. As shown in Fig. \ref{figc2},  the beam is subjected to an end load imposed on the middle point of right boundary of the design domain. These domain and boundary changes aim to test the transferability of our FE-CNN approach. Table \ref{tab3} summarizes for different resolutions of FEA meshes the optimized design ($1200\times600$ elements), the iteration count, the volume fraction, the projection radius, and the average time for the key steps of optimization process.

\begin{table}[htbp]
	\centering
	\caption{Optimization results of the cantilever beam example obtained by 88-lines code and the proposed approach with different FE meshes}
	\begin{tabular}{|c|c|c|c|c|c|c|c|c|c|}
		\hline
		\makecell{Optimized structure\\1200x600} & \makecell{Number of\\EF mesh}  & \makecell{$\bar{t}_{FEA}$\\$(s)$} 
		& \makecell{$\bar{t}_{sen}$\\$(s)$} & \makecell{$\bar{t}_{CNN}$\\$(s)$} & \makecell{$\bar{t}_{total}$\\$(s)$} 
		& \makecell{$n_{iter}$} & \makecell{$r$} & \makecell{$vf$} & \makecell{$C$}  \\
		\hline
		\begin{minipage}{0.01\textwidth}	
			\centerline{\includegraphics[width=3.0cm]{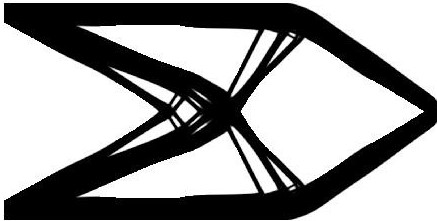}}
		\end{minipage}
		& 1200x600 & 9.57  & 0.15  &       & 10.39  & 495   & 4.0   & 0.4   & 77.03 \\
		\hline
		\begin{minipage}{0.01\textwidth}	
			\centerline{\includegraphics[width=3.0cm]{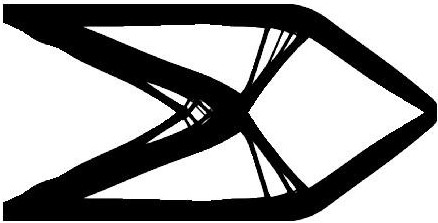}}
		\end{minipage}
		& 600x300 & 1.89  & 0.03  & 0.18  & 2.63  & 709   & 4.0   & 0.4   & 77.55 \\
		\hline
		\begin{minipage}{0.01\textwidth}	
			\centerline{\includegraphics[width=3.0cm]{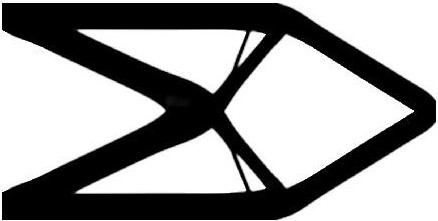}}
		\end{minipage}
		& 300x150 & 0.37  & 0.01  & 0.16  & 1.09  & 809   & 4.0   & 0.4   & 76.80 \\
		\hline
		\begin{minipage}{0.01\textwidth}	
			\centerline{\includegraphics[width=3.0cm]{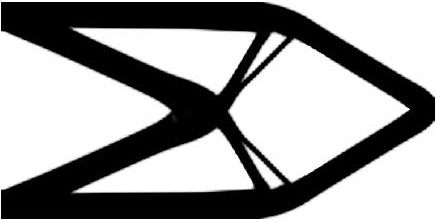}}
		\end{minipage}
		& 150x75 & 0.08  & 0.01   & 0.16  & 0.81   & 592   & 6.0   & 0.4   & 77.40 \\
		\hline
	\end{tabular}%
	\label{tab3}%
\end{table}%

\begin{figure}
	\centering
	\includegraphics[scale=1]{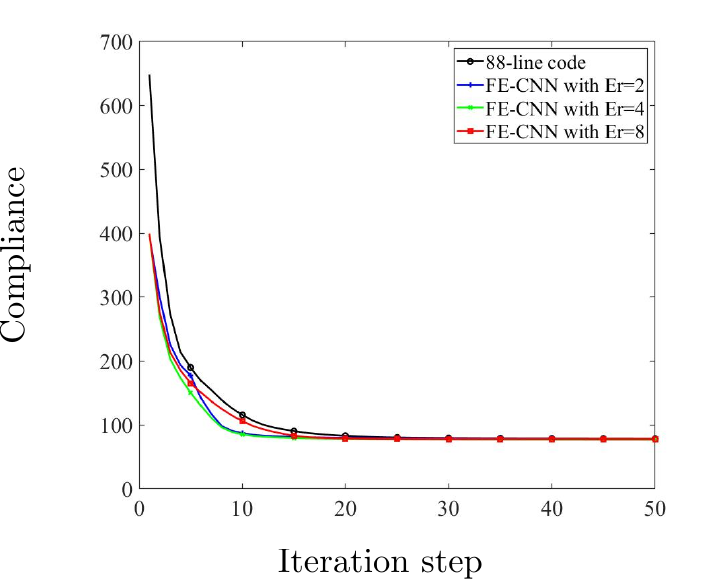}
	\caption{Convergence history after 50 iterations with design variable elements $1200\times600$.}
	\label{figl2}
\end{figure}

As seen from Table \ref{tab3}, our proposed approach requires in an iteration only about $25.6\%$, $10.5\%$ and $7.8\%$ of the average time of a corresponding iteration in the 88-line SIMP code, for scaling factors of $Er=2$, $Er=4$ and $Er=8$, respectively. The relative errors of the compliances optimized by the 88-lines of SIMP code and our FE-CNN are $0.67\%$, $0.30\%$ and $0.48\%$, respectively. The convergence histories for $50$ iterations are compared in Fig. \ref{figl2}. These optimization results confirm that  FE-CNN, trained by square images, remains effective in rectangular design domains. 

\begin{figure}
	\centering
	\includegraphics[scale=0.750]{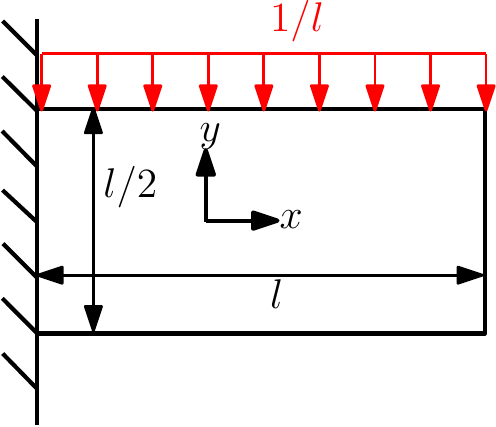}
	\caption{A cantilever beam example under uniformly distributed load}
	\label{figc3}
\end{figure}

 \subsection{Example 3: A cantilever beam subject to a distributed load; comparing to MTOP} \label{MTOPCom1}
In this and the next section the results generated through FE-CNN and MTOP, with different projection radii, are compared. First, a cantilever beam subjected to a uniformly distributed load, as introduced in \citep{groen2017higher}, is revisited. The setting of this problem is described schematically in Fig. \ref{figc3}. A vertical distributed load is imposed uniformly on the top boundary of the design domain, with a density of $1/l$. The number of design variable elements is chosen as $1200\times600$, with two scaling factors, $Er=4$ and $Er=8$, which correspond to FE meshes $300\times150$, and $150\times75$, respectively. Both FE-CNN and MTOP are tested with these same parameters, to highlight the advantage of our proposed approach. Results for scaling factors $Er=4$ and $Er=8$ are presented in Table \ref{tab4} and Table \ref{tab5}, respectively. Compared with MTOP, our FE-CNN approach can generate better structures with smaller projection radii, such as $r=1.2$ when $Er=4$, and $r=5$ when $Er=8$. In fact, for projection radii smaller than the scaling factor, there appear checkerboard patterns and discontinuities in the results of MTOP, while the results of our FE-CNN approach do not exhibit any. As noted in Section \ref{sec3.3}, the stiffness matrices and the sensitivities are computed with more descriptive mappings in FE-CNN than in MTOP. These richer mappings not only use of the results of FEA more comprehensively, but also with higher accuracy. Thus, as seen in Table \ref{tab4}, no checkerboard patterns or discontinuities appear in our results for the smaller projection radii.
Nevertheless, for a scaling factor $Er=8$ and a small projection radius, optimization results obtained by FE-CNN, like MTOP, do involve some gray-scale, blurred details (not sharply defined).

 \begin{table}[htbp]
 	\centering
 	\caption{Optimization results of the cantilever beam example obtained by FE-CNN and MTOP with scaling factor $Er=4$ (design variable elements $1200\times600$ and FE mesh $300\times150$)}
 	\begin{tabular}{|c|c|c|c|c|c|}
 		\hline
 		\makecell{MTOP\\Optimized structure} & \makecell{r}  & \makecell{detail} 
 		& \makecell{FE-CNN\\Optimized structure} & \makecell{r} & \makecell{detail}  \\
 		\hline
 		\begin{minipage}[b]{0.25\columnwidth}
 			\centering	
 			\raisebox{-.5\height}{\includegraphics[width=4.0cm]{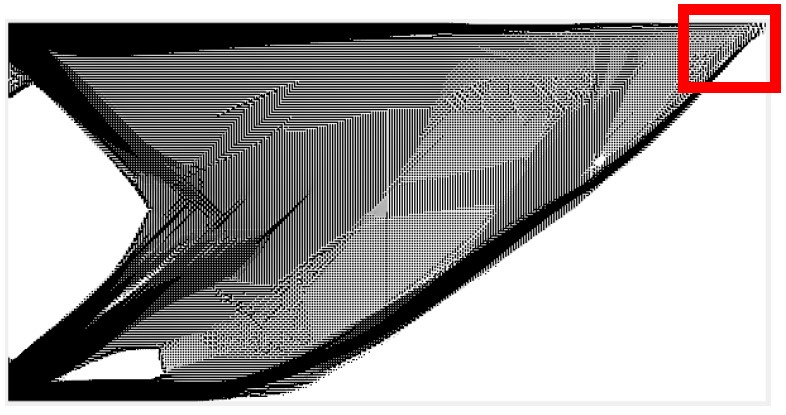}}c
 		\end{minipage}
 		& 1.2 & 
 		\begin{minipage}[b]{0.125\columnwidth}
			\centering	
 			\raisebox{-.5\height}{\includegraphics[width=2.0cm]{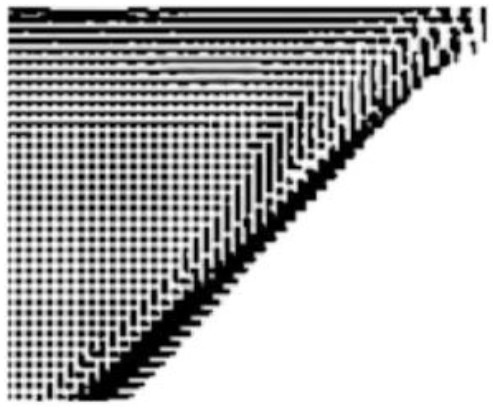}}
 		\end{minipage} &
 	 	\begin{minipage}[b]{0.25\columnwidth}
 	 		\centering	
 			\raisebox{-.5\height}{\includegraphics[width=4.0cm]{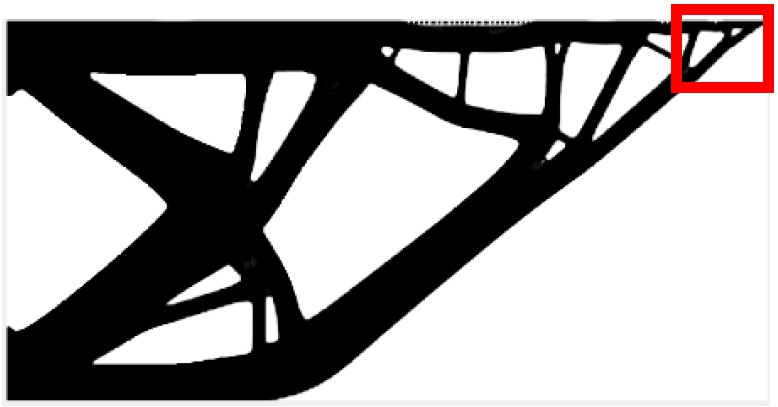}}
 		\end{minipage}
 		& 1.2 & 
 		\begin{minipage}[b]{0.125\columnwidth}
 			\centering	
 			\raisebox{-.5\height}{\includegraphics[width=2.0cm]{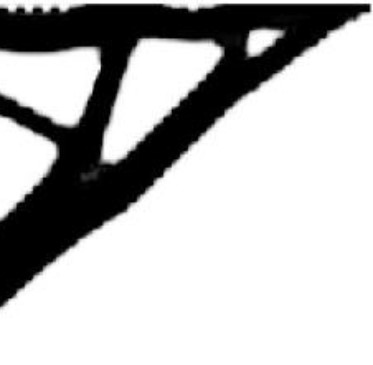}}
 		\end{minipage}\\
 		\hline
 		\begin{minipage}[b]{0.25\columnwidth}
 			\centering	
 			\raisebox{-.5\height}{\includegraphics[width=4.0cm]{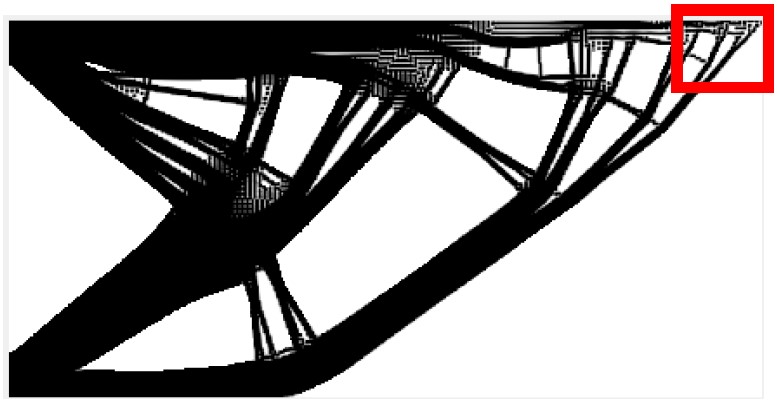}}
 		\end{minipage}
 		& 3 & 
 		\begin{minipage}[b]{0.125\columnwidth}
 			\centering	
 			\raisebox{-.5\height}{\includegraphics[width=2.0cm]{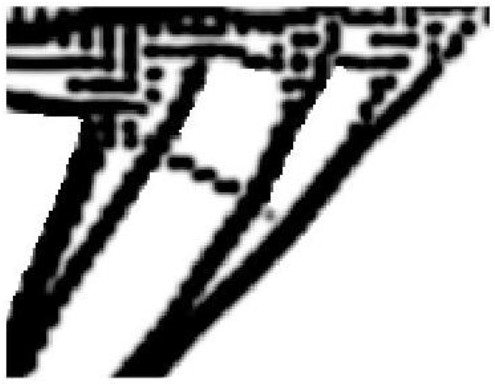}}
 		\end{minipage} &
 		\begin{minipage}[b]{0.25\columnwidth}
 			\centering	
 			\raisebox{-.5\height}{\includegraphics[width=4.0cm]{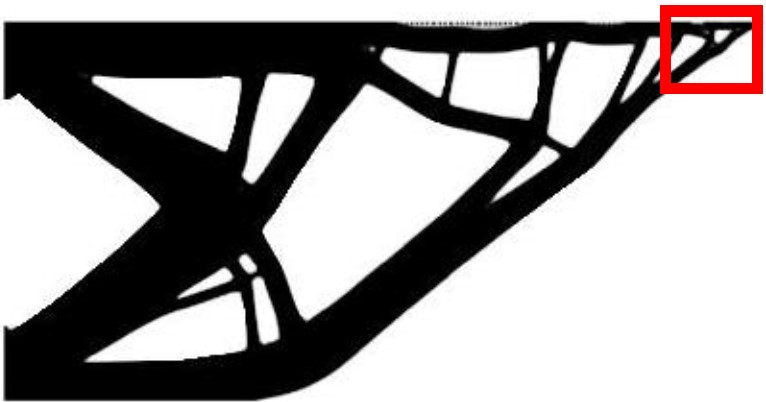}}
 		\end{minipage}
 		& 3 & 
 		\begin{minipage}[b]{0.125\columnwidth}
 			\centering	
 			\raisebox{-.5\height}{\includegraphics[width=2.0cm]{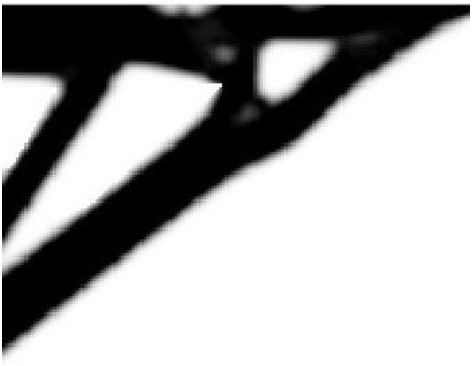}}
 		\end{minipage}\\
 		\hline
 		\begin{minipage}[b]{0.25\columnwidth}
 			\centering	
			\raisebox{-.5\height}{\includegraphics[width=4.0cm]{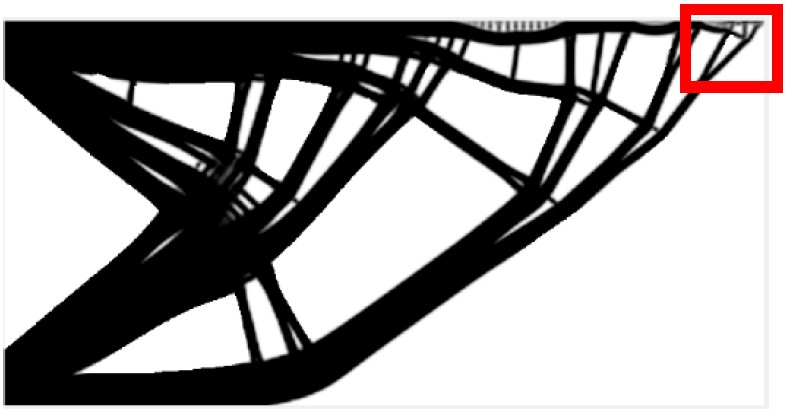}}
		\end{minipage}
		& 5 & 
		\begin{minipage}[b]{0.125\columnwidth}
			\centering	
			\raisebox{-.5\height}{\includegraphics[width=2.0cm]{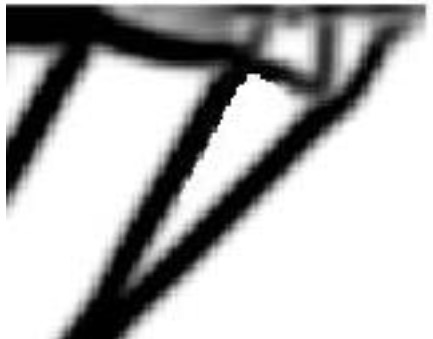}}
		\end{minipage} &
		\begin{minipage}[b]{0.25\columnwidth}
			\centering	
			\raisebox{-.5\height}{\includegraphics[width=4.0cm]{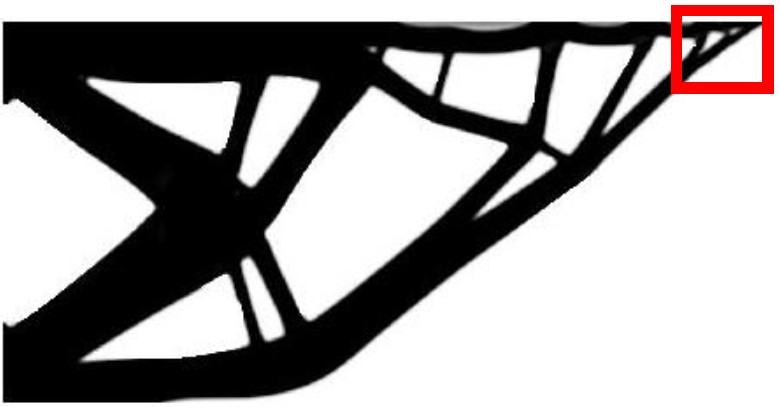}}
		\end{minipage}
		& 5 & 
		\begin{minipage}[b]{0.125\columnwidth}
			\centering	
			\raisebox{-.5\height}{\includegraphics[width=2.0cm]{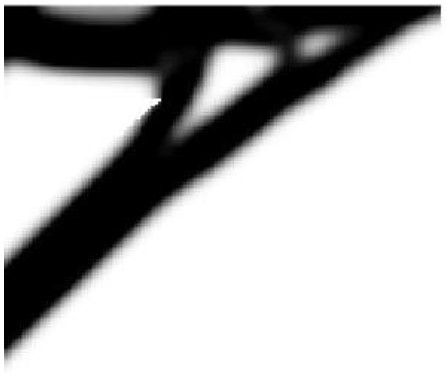}}
		\end{minipage}\\
		\hline
 	\end{tabular}%
 	\label{tab4}%
 \end{table}%

\begin{table}[htbp]
	\centering
	\caption{Optimization results of the cantilever beam example obtained by FE-CNN and MTOP with scaling factor $Er=8$ (design variable elements $1200\times600$ and FE mesh $150\times75$)}
	\begin{tabular}{|c|c|c|c|c|c|}
		\hline
		\makecell{MTOP\\Optimized structure} & \makecell{r}  & \makecell{detail} 
		& \makecell{FE-CNN\\Optimized structure} & \makecell{r} & \makecell{detail}  \\
		\hline
		\begin{minipage}[b]{0.25\columnwidth}
			\centering	
			\raisebox{-.5\height}{\includegraphics[width=4.0cm]{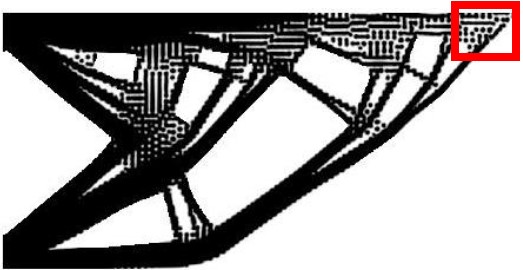}}
		\end{minipage}
		& 5 & 
		\begin{minipage}[b]{0.125\columnwidth}
			\centering	
			\raisebox{-.5\height}{\includegraphics[width=2.0cm]{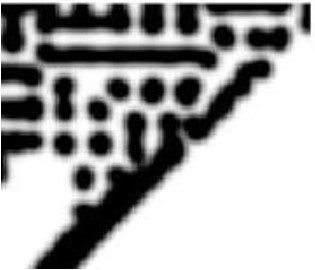}}
		\end{minipage} &
		\begin{minipage}[b]{0.25\columnwidth}
			\centering	
			\raisebox{-.5\height}{\includegraphics[width=4.0cm]{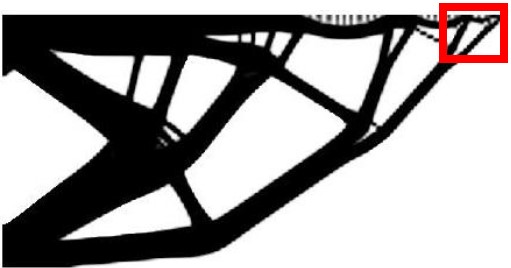}}
		\end{minipage}
		& 5 & 
		\begin{minipage}[b]{0.125\columnwidth}
			\centering	
			\raisebox{-.5\height}{\includegraphics[width=2.0cm]{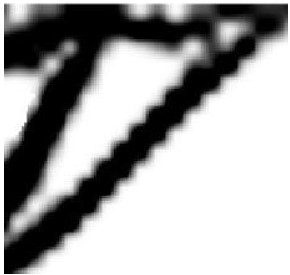}}
		\end{minipage}\\
		\hline
		\begin{minipage}[b]{0.25\columnwidth}
			\centering	
			\raisebox{-.5\height}{\includegraphics[width=4.0cm]{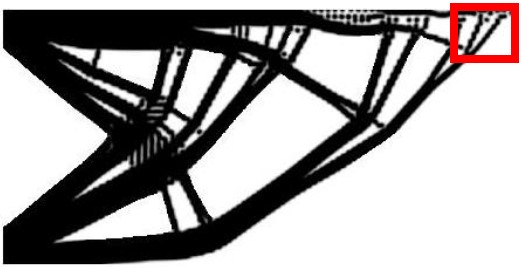}}
		\end{minipage}
		& 7 & 
		\begin{minipage}[b]{0.125\columnwidth}
			\centering	
			\raisebox{-.5\height}{\includegraphics[width=2.0cm]{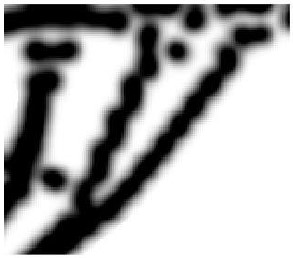}}
		\end{minipage} &
		\begin{minipage}[b]{0.25\columnwidth}
			\centering	
			\raisebox{-.5\height}{\includegraphics[width=4.0cm]{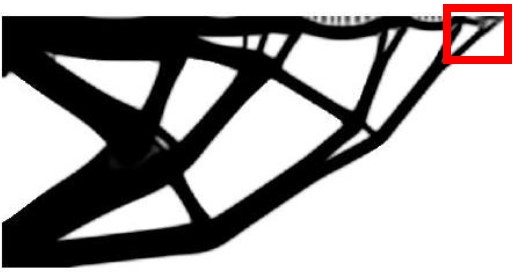}}
		\end{minipage}
		& 7 & 
		\begin{minipage}[b]{0.125\columnwidth}
			\centering	
			\raisebox{-.5\height}{\includegraphics[width=2.0cm]{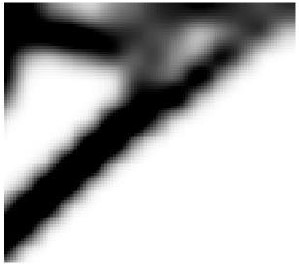}}
		\end{minipage}\\
		\hline
		\begin{minipage}[b]{0.25\columnwidth}
			\centering	
			\raisebox{-.5\height}{\includegraphics[width=4.0cm]{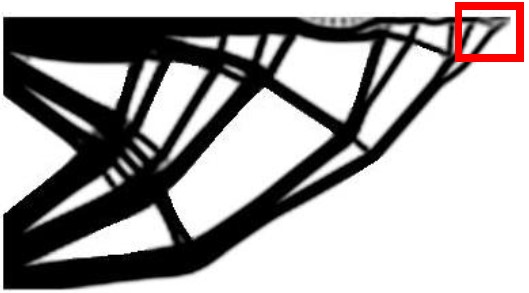}}
		\end{minipage}
		& 10 & 
		\begin{minipage}[b]{0.125\columnwidth}
			\centering	
			\raisebox{-.5\height}{\includegraphics[width=2.0cm]{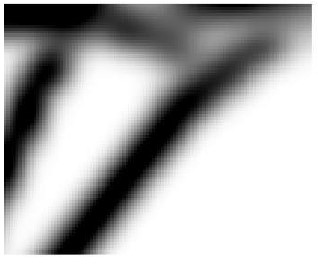}}
		\end{minipage} &
		\begin{minipage}[b]{0.25\columnwidth}
			\centering	
			\raisebox{-.5\height}{\includegraphics[width=4.0cm]{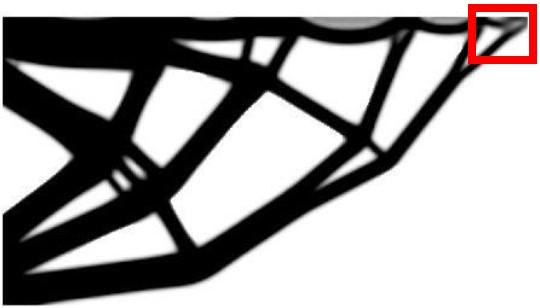}}
		\end{minipage}
		& 10 & 
		\begin{minipage}[b]{0.125\columnwidth}
			\centering	
			\raisebox{-.5\height}{\includegraphics[width=2.0cm]{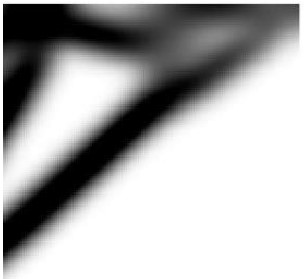}}
		\end{minipage}\\
		\hline
	\end{tabular}%
	\label{tab5}%
\end{table}%
 
 \begin{figure}
 	\centering
 	\includegraphics[scale=0.750]{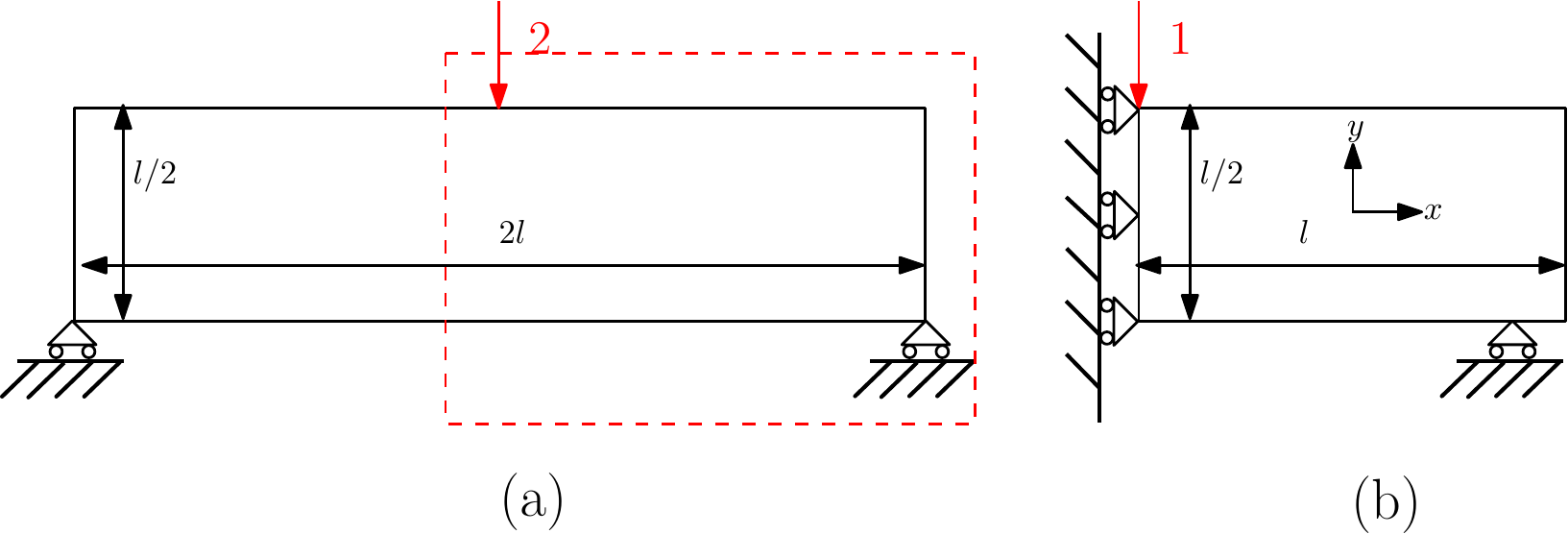}
 	\caption{The MBB example}
 	\label{figc4}
 \end{figure}

 \subsection{Example 4: An MBB beam, comparing with MTOP} \label{MTOPCom2}
We here study topology optimization of an MBB (Messerschmitt-Bolkow-Blohm) beam, to confirm the advantage of our FE-CNN approach, comparing to MTOP. The setting for this example is described schematically in Fig. \ref{figc4}. A vertical load is imposed on the middle point of the top side of the beam. For simplicity, we consider only half of the design domain, with an aspect ratio of $2:1$, and $1200\times600$ design variables. Two scaling factors $Er=4$ and $Er=8$ are picked, with corresponding FE meshes of $300\times150$ and $150\times75$, respectively. These parameters apply to both FE-CNN and MTOP. Results for both scaling factors are summarized in Table \ref{tab6} and Table \ref{tab7}, respectively. Again, similar to the results shown in Section \ref{MTOPCom1}, the optimized structures obtained through FE-CNN are acceptable, and without checkerboard patterns or discontinuities that affect MTOP with small projection radii.

\begin{table}[htbp]
	\centering
	\caption{Optimization results of the MBB example obtained by FE-CNN and MTOP with scaling factor $Er=4$ (design variable elements $1200\times600$ and FE mesh $300\times150$)}
	\begin{tabular}{|c|c|c|c|c|c|}
		\hline
		\makecell{MTOP\\Optimized structure} & \makecell{r}  & \makecell{detail} 
		& \makecell{FE-CNN\\Optimized structure} & \makecell{r} & \makecell{detail}  \\
		\hline
		\begin{minipage}[b]{0.25\columnwidth}
			\centering	
			\raisebox{-.5\height}{\includegraphics[width=4.0cm]{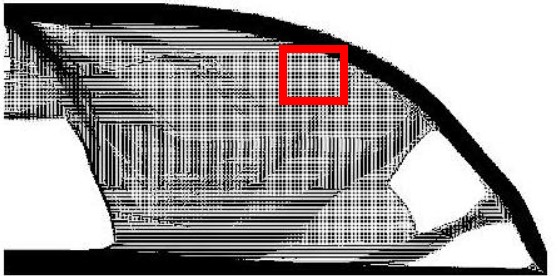}}
		\end{minipage}
		& 1.2 & 
		\begin{minipage}[b]{0.125\columnwidth}
			\centering	
			\raisebox{-.5\height}{\includegraphics[width=2.0cm]{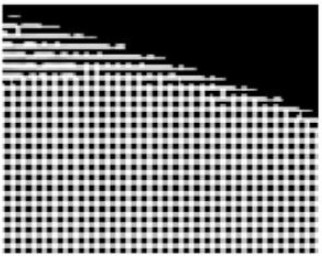}}
		\end{minipage} &
		\begin{minipage}[b]{0.25\columnwidth}
			\centering	
			\raisebox{-.5\height}{\includegraphics[width=4.0cm]{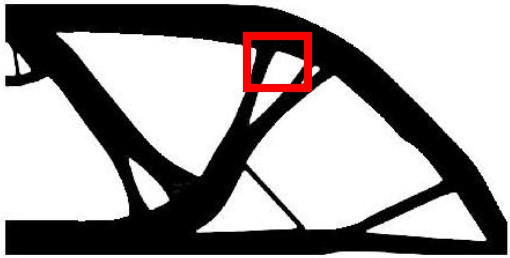}}
		\end{minipage}
		& 1.2 & 
		\begin{minipage}[b]{0.125\columnwidth}
			\centering	
			\raisebox{-.5\height}{\includegraphics[width=2.0cm]{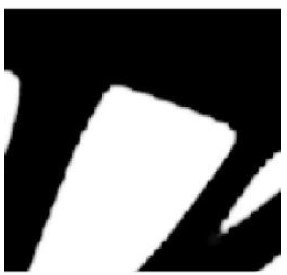}}
		\end{minipage}\\
		\hline
		\begin{minipage}[b]{0.25\columnwidth}
			\centering	
			\raisebox{-.5\height}{\includegraphics[width=4.0cm]{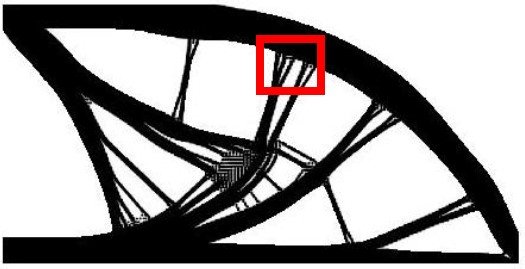}}
		\end{minipage}
		& 3 & 
		\begin{minipage}[b]{0.125\columnwidth}
			\centering	
			\raisebox{-.5\height}{\includegraphics[width=2.0cm]{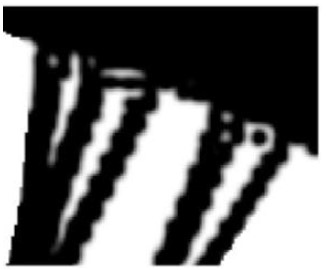}}
		\end{minipage} &
		\begin{minipage}[b]{0.25\columnwidth}
			\centering	
			\raisebox{-.5\height}{\includegraphics[width=4.0cm]{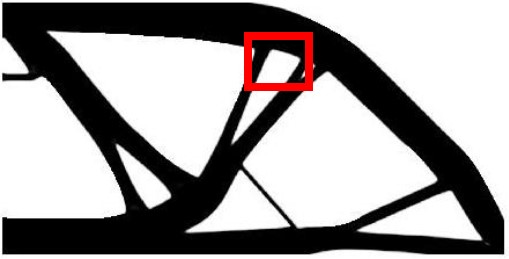}}
		\end{minipage}
		& 3 & 
		\begin{minipage}[b]{0.125\columnwidth}
			\centering	
			\raisebox{-.5\height}{\includegraphics[width=2.0cm]{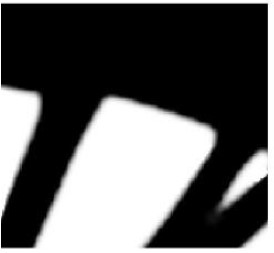}}
		\end{minipage}\\
		\hline
		\begin{minipage}[b]{0.25\columnwidth}
			\centering	
			\raisebox{-.5\height}{\includegraphics[width=4.0cm]{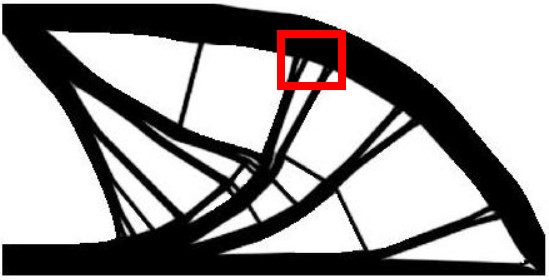}}
		\end{minipage}
		& 5 & 
		\begin{minipage}[b]{0.125\columnwidth}
			\centering	
			\raisebox{-.5\height}{\includegraphics[width=2.0cm]{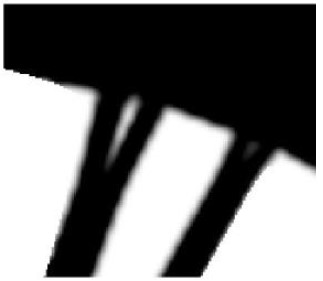}}
		\end{minipage} &
		\begin{minipage}[b]{0.25\columnwidth}
			\centering	
			\raisebox{-.5\height}{\includegraphics[width=4.0cm]{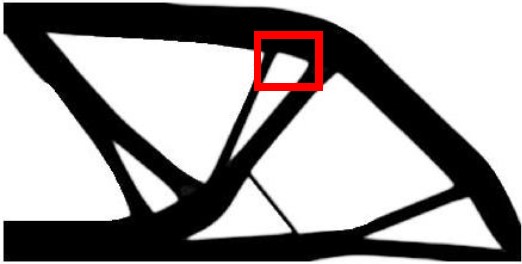}}
		\end{minipage}
		& 5 & 
		\begin{minipage}[b]{0.125\columnwidth}
			\centering	
			\raisebox{-.5\height}{\includegraphics[width=2.0cm]{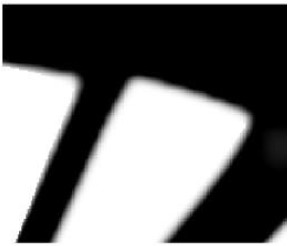}}
		\end{minipage}\\
		\hline
	\end{tabular}%
	\label{tab6}%
\end{table}%

\begin{table}[htbp]
	\centering
	\caption{Optimization results of the MBB example obtained by FE-CNN and MTOP with scaling factor $Er=8$ (design variable elements $1200\times600$ and FE mesh $150\times75$)}
	\begin{tabular}{|c|c|c|c|c|c|}
		\hline
		\makecell{MTOP\\Optimized structure} & \makecell{r}  & \makecell{detail} 
		& \makecell{FE-CNN\\Optimized structure} & \makecell{r} & \makecell{detail}  \\
		\hline
	\begin{minipage}[b]{0.25\columnwidth}
		\centering	
		\raisebox{-.5\height}{\includegraphics[width=4.0cm]{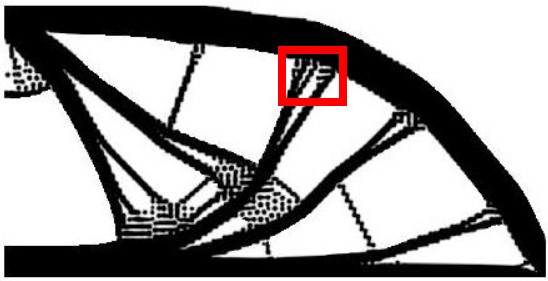}}
	\end{minipage}
	& 5 & 
	\begin{minipage}[b]{0.125\columnwidth}
		\centering	
		\raisebox{-.5\height}{\includegraphics[width=2.0cm]{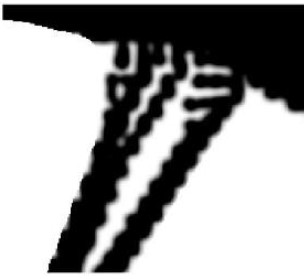}}
	\end{minipage} &
	\begin{minipage}[b]{0.25\columnwidth}
		\centering	
		\raisebox{-.5\height}{\includegraphics[width=4.0cm]{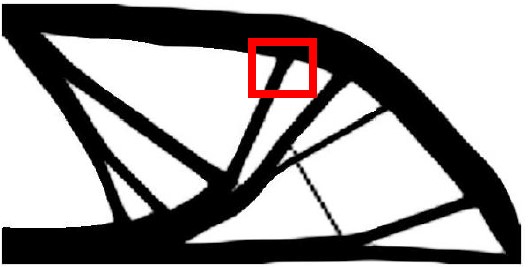}}
	\end{minipage}
	& 5 & 
	\begin{minipage}[b]{0.125\columnwidth}
		\centering	
		\raisebox{-.5\height}{\includegraphics[width=2.0cm]{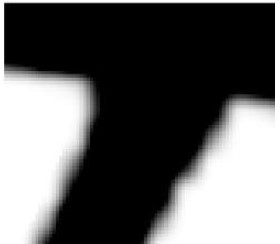}}
	\end{minipage}\\
	\hline
	\begin{minipage}[b]{0.25\columnwidth}
		\centering	
		\raisebox{-.5\height}{\includegraphics[width=4.0cm]{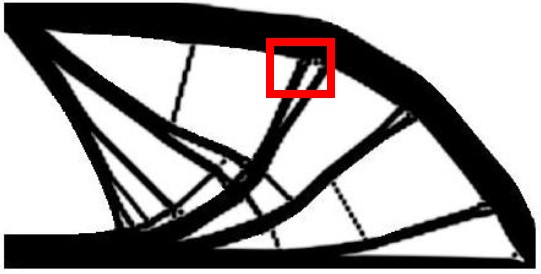}}
	\end{minipage}
	& 7 & 
	\begin{minipage}[b]{0.125\columnwidth}
		\centering	
		\raisebox{-.5\height}{\includegraphics[width=2.0cm]{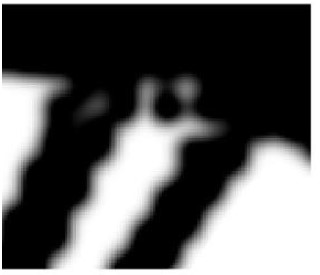}}
	\end{minipage} &
	\begin{minipage}[b]{0.25\columnwidth}
		\centering	
		\raisebox{-.5\height}{\includegraphics[width=4.0cm]{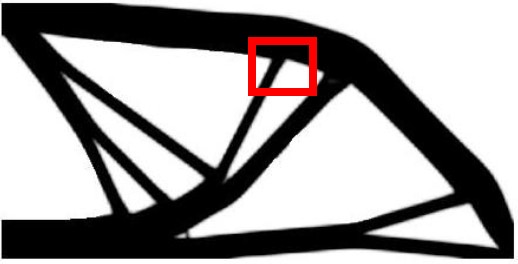}}
	\end{minipage}
	& 7 & 
	\begin{minipage}[b]{0.125\columnwidth}
		\centering	
		\raisebox{-.5\height}{\includegraphics[width=2.0cm]{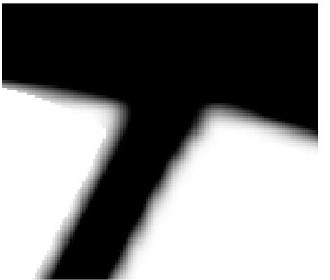}}
	\end{minipage}\\
	\hline
	\begin{minipage}[b]{0.25\columnwidth}
		\centering	
		\raisebox{-.5\height}{\includegraphics[width=4.0cm]{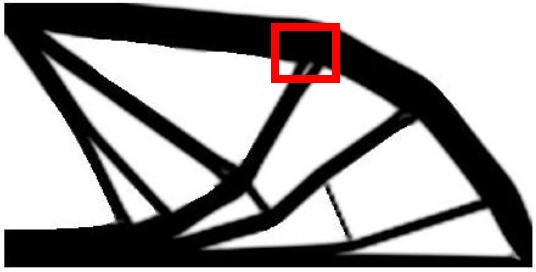}}
	\end{minipage}
	& 10 & 
	\begin{minipage}[b]{0.125\columnwidth}
		\centering	
		\raisebox{-.5\height}{\includegraphics[width=2.0cm]{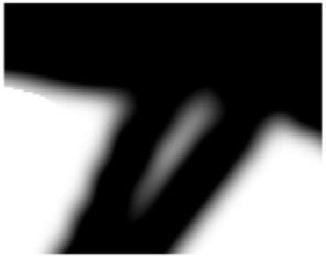}}
	\end{minipage} &
	\begin{minipage}[b]{0.25\columnwidth}
		\centering	
		\raisebox{-.5\height}{\includegraphics[width=4.0cm]{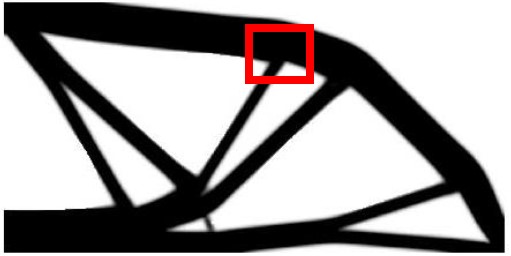}}
	\end{minipage}
	& 10 & 
	\begin{minipage}[b]{0.125\columnwidth}
		\centering	
		\raisebox{-.5\height}{\includegraphics[width=2.0cm]{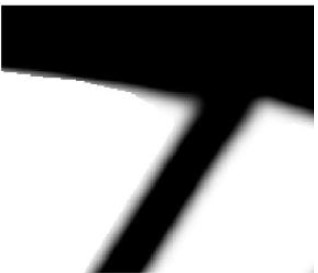}}
	\end{minipage}\\
	\hline
	\end{tabular}%
		\label{tab7}%
\end{table}%

\section{Conclusion}
In this paper, a data-driven approach is proposed to accelerate  structural topology optimization, by means of an in-house developed finite element convolution neural network (FE-CNN). An OC method, followed by FEA, are used to replace the activation function and the fully connected layer in conventional CNNs. The unpooling and BP-Conv layers are herein used as decoders. With this approach, the design variables can be greatly reduced in large-scale topology optimization problems. The FE-CNN is trained by using a small-scale image. After training, the sensitivities of the compliance for  large-scale topology optimization can be computed from the small-scale meshes. Various numerical examples were performed to demonstrate the validity of our approach, and its advantages when compared to an efficient implementation of SIMP and of MTOP. All  results showed sufficient accuracy for our FE-CNN method, and confirmed that the computational cost is reduced by up to an order of magnitude; an essential feature of our proposed approach to countering the curse of dimensionality. Compared with MTOP, our proposed approach can compute more accurate and smoother sensitivities, and generate better optimized results. As an initial implementation, only compliance was considered for the objective of offline training in this present study. Other objective functions, such as maximizing the fundamental frequency, can be investigated. Similar avenues motivate future investigations into the development of FE-CNN.

\section{Acknowledgment}
Z. D. acknowledges the support from NSF of China (12002073) and Fundamental research founds for the central universities of China (DUT20RC(3)020). 
C. L. acknowledges the support from NSF of China (12002077) and China Postdoctoral Science Foundation funded project (2020T130078, 2020M680944) 
S.T. appreciates the support from NSF of China (Project No. 11872139). 
X.G. Thanks the support from NSF of China (11732004, 11821202), and Program for Changjiang Scholars, Innovative Research Team in University (PCSIRT).


\bibliographystyle{elsarticle-num-names}
\biboptions{square,numbers,sort&compress}
\bibliography{mybibfile.bib}

\begin{thebibliography}{27}
\expandafter\ifx\csname natexlab\endcsname\relax\def\natexlab#1{#1}\fi
\providecommand{\url}[1]{\texttt{#1}}
\providecommand{\href}[2]{#2}
\providecommand{\path}[1]{#1}
\providecommand{\DOIprefix}{doi:}
\providecommand{\ArXivprefix}{arXiv:}
\providecommand{\URLprefix}{URL: }
\providecommand{\Pubmedprefix}{pmid:}
\providecommand{\doi}[1]{\href{http://dx.doi.org/#1}{\path{#1}}}
\providecommand{\Pubmed}[1]{\href{pmid:#1}{\path{#1}}}
\providecommand{\bibinfo}[2]{#2}
\ifx\xfnm\relax \def\xfnm[#1]{\unskip,\space#1}\fi
\bibitem[{Prager and Rozvany(1977)}]{prager1977optimal}
\bibinfo{author}{W.~Prager}, \bibinfo{author}{G.~Rozvany},
\newblock \bibinfo{title}{Optimal layout of grillages},
\newblock \bibinfo{journal}{Journal of Structural Mechanics}
  \bibinfo{volume}{5} (\bibinfo{year}{1977}) \bibinfo{pages}{1--18}.
  \DOIprefix\doi{10.1080/03601217708907301}.
\bibitem[{Cheng and Olhoff(1981)}]{cheng1981investigation}
\bibinfo{author}{K.-T. Cheng}, \bibinfo{author}{N.~Olhoff},
\newblock \bibinfo{title}{An investigation concerning optimal design of solid
  elastic plates},
\newblock \bibinfo{journal}{International Journal of Solids and Structures}
  \bibinfo{volume}{17} (\bibinfo{year}{1981}) \bibinfo{pages}{305--323}.
  \DOIprefix\doi{10.1016/0020-7683(81)90065-2}.
\bibitem[{Bendsøe and Kikuchi(1988)}]{bendsoe1988generating}
\bibinfo{author}{M.~P. Bendsøe}, \bibinfo{author}{N.~Kikuchi},
\newblock \bibinfo{title}{Generating optimal topologies in structural design
  using a homogenization method},
\newblock \bibinfo{journal}{Computer Methods in Applied Mechanics and
  Engineering} \bibinfo{volume}{71(2)} (\bibinfo{year}{1988})
  \bibinfo{pages}{197–224}. \DOIprefix\doi{10.1016/0045-7825(88)90086-2}.
\bibitem[{Zhou and Rozvany(1991)}]{zhou1991coc}
\bibinfo{author}{M.~Zhou}, \bibinfo{author}{G.~Rozvany},
\newblock \bibinfo{title}{The coc algorithm, part ii: Topological, geometrical
  and generalized shape optimization},
\newblock \bibinfo{journal}{Computer Methods in Applied Mechanics and
  Engineering} \bibinfo{volume}{89} (\bibinfo{year}{1991})
  \bibinfo{pages}{309--336}. \DOIprefix\doi{10.1016/0045-7825(91)90046-9}.
\bibitem[{Rozvany(2009)}]{rozvany2009critical}
\bibinfo{author}{G.~I. Rozvany},
\newblock \bibinfo{title}{A critical review of established methods of
  structural topology optimization},
\newblock \bibinfo{journal}{Structural and Multidisciplinary Optimization}
  \bibinfo{volume}{37} (\bibinfo{year}{2009}) \bibinfo{pages}{217--237}.
  \DOIprefix\doi{10.1007/s00158-007-0217-0}.
\bibitem[{Guo and Cheng(2010)}]{guo2010recent}
\bibinfo{author}{X.~Guo}, \bibinfo{author}{G.-D. Cheng},
\newblock \bibinfo{title}{Recent development in structural design and
  optimization},
\newblock \bibinfo{journal}{Acta Mechanica Sinica} \bibinfo{volume}{26}
  (\bibinfo{year}{2010}) \bibinfo{pages}{807--823}.
  \DOIprefix\doi{10.1007/s10409-010-0395-7}.
\bibitem[{Maute and Sigmund(2013)}]{maute2013topology}
\bibinfo{author}{K.~Maute}, \bibinfo{author}{O.~Sigmund},
\newblock \bibinfo{title}{Topology optimization approaches: A comparative
  review},
\newblock \bibinfo{journal}{Structural and Multidisciplinary Optimization}
  \bibinfo{volume}{48} (\bibinfo{year}{2013}) \bibinfo{pages}{1031–1055}.
  \DOIprefix\doi{10.1007/s00158-013-0978-6}.
\bibitem[{Deaton and Grandhi(2014)}]{deaton2014survey}
\bibinfo{author}{J.~D. Deaton}, \bibinfo{author}{R.~V. Grandhi},
\newblock \bibinfo{title}{A survey of structural and multidisciplinary
  continuum topology optimization: post 2000},
\newblock \bibinfo{journal}{Structural and Multidisciplinary Optimization}
  \bibinfo{volume}{49} (\bibinfo{year}{2014}) \bibinfo{pages}{1--38}.
  \DOIprefix\doi{10.1007/s00158-013-0956-z}.
\bibitem[{Bends{\o}e(1989)}]{bendsoe1989optimal}
\bibinfo{author}{M.~P. Bends{\o}e},
\newblock \bibinfo{title}{Optimal shape design as a material distribution
  problem},
\newblock \bibinfo{journal}{Structural optimization} \bibinfo{volume}{1}
  (\bibinfo{year}{1989}) \bibinfo{pages}{193--202}.
\bibitem[{Rozvany et~al.(1989)Rozvany, Zhou, Rotthaus, Gollub, and
  Spengemann}]{rozvany1989continuum}
\bibinfo{author}{G.~Rozvany}, \bibinfo{author}{M.~Zhou},
  \bibinfo{author}{M.~Rotthaus}, \bibinfo{author}{W.~Gollub},
  \bibinfo{author}{F.~Spengemann},
\newblock \bibinfo{title}{Continuum-type optimality criteria methods for large
  finite element systems with a displacement constraint. part i},
\newblock \bibinfo{journal}{Structural optimization} \bibinfo{volume}{1}
  (\bibinfo{year}{1989}) \bibinfo{pages}{47--72}.
\bibitem[{Ulu et~al.(2016)Ulu, Zhang, and Kara}]{ulu2016data}
\bibinfo{author}{E.~Ulu}, \bibinfo{author}{R.~Zhang}, \bibinfo{author}{L.~B.
  Kara},
\newblock \bibinfo{title}{A data-driven investigation and estimation of optimal
  topologies under variable loading configurations},
\newblock \bibinfo{journal}{Computer Methods in Biomechanics and Biomedical
  Engineering: Imaging \& Visualization} \bibinfo{volume}{4}
  (\bibinfo{year}{2016}) \bibinfo{pages}{61--72}.
\bibitem[{Sosnovik and Oseledets(2019)}]{sosnovik2019neural}
\bibinfo{author}{I.~Sosnovik}, \bibinfo{author}{I.~Oseledets},
\newblock \bibinfo{title}{Neural networks for topology optimization},
\newblock \bibinfo{journal}{Russian Journal of Numerical Analysis and
  Mathematical Modelling} \bibinfo{volume}{34} (\bibinfo{year}{2019})
  \bibinfo{pages}{215--223}.
\bibitem[{Banga et~al.(2018)Banga, Gehani, Bhilare, Patel, and
  Kara}]{banga20183d}
\bibinfo{author}{S.~Banga}, \bibinfo{author}{H.~Gehani},
  \bibinfo{author}{S.~Bhilare}, \bibinfo{author}{S.~Patel},
  \bibinfo{author}{L.~Kara},
\newblock \bibinfo{title}{3d topology optimization using convolutional neural
  networks},
\newblock \bibinfo{journal}{arXiv preprint: 1808.07440}
  (\bibinfo{year}{2018}).
\bibitem[{Yu et~al.(2019)Yu, Hur, Jung, and Jang}]{Yu}
\bibinfo{author}{Y.~Yu}, \bibinfo{author}{T.~Hur}, \bibinfo{author}{J.~Jung},
  \bibinfo{author}{I.~G. Jang},
\newblock \bibinfo{title}{Deep learning for determining a near-optimal
  topological design without any iteration},
\newblock \bibinfo{journal}{Structural and Multidisciplinary Optimization}
  \bibinfo{volume}{59} (\bibinfo{year}{2019}) \bibinfo{pages}{787--799}.
\bibitem[{Lee et~al.(2020)Lee, Kim, Lieu, and Lee}]{Lee}
\bibinfo{author}{S.~Lee}, \bibinfo{author}{H.~Kim}, \bibinfo{author}{Q.~X.
  Lieu}, \bibinfo{author}{J.~Lee},
\newblock \bibinfo{title}{Cnn-based image recognition for topology
  optimization},
\newblock \bibinfo{journal}{Knowledge-Based Systems}  (\bibinfo{year}{2020})
  \bibinfo{pages}{105887}.
\bibitem[{Takahashi et~al.(2019)Takahashi, Suzuki, and Todoroki}]{Takahashi}
\bibinfo{author}{Y.~Takahashi}, \bibinfo{author}{Y.~Suzuki},
  \bibinfo{author}{A.~Todoroki},
\newblock \bibinfo{title}{Convolutional neural network-based topology
  optimization (cnn-to) by estimating sensitivity of compliance from material
  distribution},
\newblock \bibinfo{journal}{arXiv preprint: 2001.00635}
  (\bibinfo{year}{2019}).
\bibitem[{Lei et~al.(2019)Lei, Liu, Du, Zhang, and Guo}]{lei2019machine}
\bibinfo{author}{X.~Lei}, \bibinfo{author}{C.~Liu}, \bibinfo{author}{Z.~Du},
  \bibinfo{author}{W.~Zhang}, \bibinfo{author}{X.~Guo},
\newblock \bibinfo{title}{Machine learning-driven real-time topology
  optimization under moving morphable component-based framework},
\newblock \bibinfo{journal}{Journal of Applied Mechanics} \bibinfo{volume}{86}
  (\bibinfo{year}{2019}).
\bibitem[{Guo et~al.(2014)Guo, Zhang, and Zhong}]{guo2014doing}
\bibinfo{author}{X.~Guo}, \bibinfo{author}{W.~Zhang},
  \bibinfo{author}{W.~Zhong},
\newblock \bibinfo{title}{Doing topology optimization explicitly and
  geometrically—a new moving morphable components based framework},
\newblock \bibinfo{journal}{Journal of Applied Mechanics} \bibinfo{volume}{81}
  (\bibinfo{year}{2014}).
\bibitem[{Guo et~al.(2016)Guo, Zhang, Zhang, and Yuan}]{guo2016explicit}
\bibinfo{author}{X.~Guo}, \bibinfo{author}{W.~Zhang},
  \bibinfo{author}{J.~Zhang}, \bibinfo{author}{J.~Yuan},
\newblock \bibinfo{title}{Explicit structural topology optimization based on
  moving morphable components (mmc) with curved skeletons},
\newblock \bibinfo{journal}{Computer methods in applied mechanics and
  engineering} \bibinfo{volume}{310} (\bibinfo{year}{2016})
  \bibinfo{pages}{711--748}.
\bibitem[{Zhang et~al.(2016)Zhang, Yuan, Zhang, and Guo}]{zhang2016new}
\bibinfo{author}{W.~Zhang}, \bibinfo{author}{J.~Yuan},
  \bibinfo{author}{J.~Zhang}, \bibinfo{author}{X.~Guo},
\newblock \bibinfo{title}{A new topology optimization approach based on moving
  morphable components (mmc) and the ersatz material model},
\newblock \bibinfo{journal}{Structural and Multidisciplinary Optimization}
  \bibinfo{volume}{53} (\bibinfo{year}{2016}) \bibinfo{pages}{1243--1260}.
\bibitem[{Chi et~al.(2021)Chi, Zhang, Tang, Mirabella, Dalloro, Song, and
  Paulino}]{chi2021universal}
\bibinfo{author}{H.~Chi}, \bibinfo{author}{Y.~Zhang}, \bibinfo{author}{T.~L.~E.
  Tang}, \bibinfo{author}{L.~Mirabella}, \bibinfo{author}{L.~Dalloro},
  \bibinfo{author}{L.~Song}, \bibinfo{author}{G.~H. Paulino},
\newblock \bibinfo{title}{Universal machine learning for topology
  optimization},
\newblock \bibinfo{journal}{Computer Methods in Applied Mechanics and
  Engineering} \bibinfo{volume}{375} (\bibinfo{year}{2021})
  \bibinfo{pages}{112739}.
\bibitem[{Nguyen et~al.(2010)Nguyen, Paulino, Song, and
  Le}]{nguyen2010computational}
\bibinfo{author}{T.~H. Nguyen}, \bibinfo{author}{G.~H. Paulino},
  \bibinfo{author}{J.~Song}, \bibinfo{author}{C.~H. Le},
\newblock \bibinfo{title}{A computational paradigm for multiresolution topology
  optimization (mtop)},
\newblock \bibinfo{journal}{Structural and Multidisciplinary Optimization}
  \bibinfo{volume}{41} (\bibinfo{year}{2010}) \bibinfo{pages}{525--539}.
\bibitem[{Liu et~al.(2018)Liu, Zhu, Sun, Li, Du, Zhang, and
  Guo}]{liu2018efficient}
\bibinfo{author}{C.~Liu}, \bibinfo{author}{Y.~Zhu}, \bibinfo{author}{Z.~Sun},
  \bibinfo{author}{D.~Li}, \bibinfo{author}{Z.~Du}, \bibinfo{author}{W.~Zhang},
  \bibinfo{author}{X.~Guo},
\newblock \bibinfo{title}{An efficient moving morphable component (mmc)-based
  approach for multi-resolution topology optimization},
\newblock \bibinfo{journal}{Structural and Multidisciplinary Optimization}
  \bibinfo{volume}{58} (\bibinfo{year}{2018}) \bibinfo{pages}{2455--2479}.
\bibitem[{Sigmund(2001)}]{sigmund200199}
\bibinfo{author}{O.~Sigmund},
\newblock \bibinfo{title}{A 99 line topology optimization code written in
  matlab},
\newblock \bibinfo{journal}{Structural and multidisciplinary optimization}
  \bibinfo{volume}{21} (\bibinfo{year}{2001}) \bibinfo{pages}{120--127}.
\bibitem[{Bouvrie(2006)}]{bouvrie2006notes}
\bibinfo{author}{J.~Bouvrie},
\newblock \bibinfo{title}{Notes on convolutional neural networks},
\newblock \bibinfo{journal}{MIT CBCL Tech Report}  (\bibinfo{year}{2006}).
\bibitem[{Andreassen et~al.(2011)Andreassen, Clausen, Schevenels, Lazarov, and
  Sigmund}]{andreassen2011efficient}
\bibinfo{author}{E.~Andreassen}, \bibinfo{author}{A.~Clausen},
  \bibinfo{author}{M.~Schevenels}, \bibinfo{author}{B.~S. Lazarov},
  \bibinfo{author}{O.~Sigmund},
\newblock \bibinfo{title}{Efficient topology optimization in matlab using 88
  lines of code},
\newblock \bibinfo{journal}{Structural and Multidisciplinary Optimization}
  \bibinfo{volume}{43} (\bibinfo{year}{2011}) \bibinfo{pages}{1--16}.
\bibitem[{Groen et~al.(2017)Groen, Langelaar, Sigmund, and
  Ruess}]{groen2017higher}
\bibinfo{author}{J.~P. Groen}, \bibinfo{author}{M.~Langelaar},
  \bibinfo{author}{O.~Sigmund}, \bibinfo{author}{M.~Ruess},
\newblock \bibinfo{title}{Higher-order multi-resolution topology optimization
  using the finite cell method},
\newblock \bibinfo{journal}{International Journal for Numerical Methods in
  Engineering} \bibinfo{volume}{110} (\bibinfo{year}{2017})
  \bibinfo{pages}{903--920}.

\end{thebibliography}

\end{document}